\documentclass{article} 
\usepackage{iclr2025_conference,times}

\usepackage{amsmath,amsfonts,bm}

\def\eqref#1{equation~\ref{#1}}

\def\1{\bm{1}}

\DeclareMathAlphabet{\mathsfit}{\encodingdefault}{\sfdefault}{m}{sl}
\SetMathAlphabet{\mathsfit}{bold}{\encodingdefault}{\sfdefault}{bx}{n}

\usepackage{hyperref}
\usepackage{url}
\usepackage{graphicx}
\usepackage{amsmath} 
\usepackage{subcaption}
\usepackage{wrapfig}
\usepackage{booktabs}
\usepackage{float}

\title{Self-Attention-Based Contextual Modulation Improves Neural System Identification}

\author{
Isaac Lin$^{1,*}$, Tianye Wang$^2$, Shang Gao$^{1,3}$, Shiming Tang$^2$, Tai Sing Lee$^{1,*}$ \\
$^1$Carnegie Mellon University, $^2$Peking University, $^3$Massachusetts Institute of Technology \\
}

\iclrfinalcopy 
\begin{document}

\renewcommand{\thefootnote}{\fnsymbol{footnote}}
\footnotetext[1]{ Corresponding authors: Isaac Lin \texttt{(isaacl@cs.cmu.edu)}, Tai Sing Lee \texttt{(taislee@cmu.edu)}.}

\maketitle

\begin{abstract}
 Convolutional neural networks (CNNs) have been shown to be state-of-the-art models for visual cortical neurons. Cortical neurons in the primary visual cortex are sensitive to contextual information mediated by extensive horizontal and feedback connections. Standard CNNs integrate global contextual information to model contextual modulation via two mechanisms: successive convolutions and a fully connected readout layer. In this paper, we find that self-attention (SA), an implementation of non-local network mechanisms, can improve neural response predictions over parameter-matched CNNs in two key metrics: tuning curve correlation and peak tuning. We introduce peak tuning as a metric to evaluate a model's ability to capture a neuron's top feature preference. We factorize networks to assess each context mechanism, revealing that information in the local receptive field is most important for modeling overall tuning, but surround information is critically necessary for characterizing the tuning peak. We find that self-attention can replace posterior spatial-integration convolutions when learned incrementally, and is further enhanced in the presence of a fully connected readout layer, suggesting that the two context mechanisms are complementary. Finally, we find that decomposing receptive field learning and contextual modulation learning in an incremental manner may be an effective and robust mechanism for learning surround-center interactions.
\end{abstract}

\section{Introduction}
Feedforward CNN models have been shown in recent years to be an effective approach for modeling and predicting visual cortical neurons' responses to arbitrary natural images \citep{10.1167/19.4.29,klindt2018neural,Yamins2016UsingGD,Zhang2018ConvolutionalNN,Cadena201764,annurev:/content/journals/10.1146/annurev-vision-082114-035447}. Neurons in the primate visual cortex are known to have extensive horizontal and feedback recurrent connections for mediating contextual modulation \citep{Felleman1991DistributedHP,https://doi.org/10.1002/cne.23458}. Feedforward CNNs can model the influence of contextual surround on the responses of the neurons via two mechanisms: successive convolution layers and a fully connected layer. Both can make the neural model's responses sensitive to the global image context, outside the traditional classical receptive fields of neurons. Broadly speaking, feedforward CNNs trained on image classification tasks have been shown to inherently exhibit contextual surround modulation similar to what has been observed in the visual cortex \cite{Pan2023.03.18.533295}.

In the context of neural prediction, it is found that including the inductive bias of horizontal recurrent connections can improve the model's predictive capabilities \citep{zhang2022recurrentnetworksimproveneural}, and that replacing a feedforward layer with a recurrent layer using a Markovian local kernel consistently outperforms parameter-matched feedforward CNNs in image classification tasks \citep{han2018deep,nayebi2018taskdriven,kubilius2019brainlike,zhang2022recurrentnetworksimproveneural}. However, contextual modulation in the visual cortex involves both the near surround and far surround, with the far surround being mediated by top-down feedback \citep{ANGELUCCI200693, Sasaki93,Shushruth2013DifferentOT}. In addition, there is evidence that contextual modulation is dynamic and highly image-dependent, suggesting a flexible gating mechanism \citep{article}. Such a flexible gating mechanism can be modeled by a combination of Gaussian mixture models, implemented either by image-dependent normalization \citep{article} or by non-local networks and the self-attention mechanism in deep learning \citep{fei2022attention}. Self-attention-based architectures such as vision transformers have recently been shown to be effective in modeling mouse V1 neurons \citep{li2023v1tlargescalemousev1}. However, these networks, often using a large number of layers and multiple attention heads, may be unnecessarily complex for this task.

In this paper, we demonstrate that a simple self-attention layer coupled with a CNN is sufficient in improving neural response prediction of macaque V1 neurons in two performance metrics: overall tuning correlation and prediction of the tuning peaks. To understand the mechanism driving improvement, we assessed the three contextual modulation mechanisms -- convolutions, self-attention, and a fully connected readout layer. We found that while the three context mechanisms complement one another to produce the best prediction performance when used in conjunction, they have specific roles. First, the fully connected layer plays a critical role in peak prediction, though self-attention can further enhance it. Second, self-attention alone can improve tuning curve correlation but is insufficient for predicting the response peak. The performance of self-attention models can be greatly enhanced when the feedforward receptive fields are learned first before learning the self-attention network, rather than learning everything simultaneously. The benefits of such incremental learning \citep{incart,cotton2020factorized} in this context are novel, suggesting that decoupling the learning of feedforward receptive fields and recurrent connections allows the system to learn a richer representation of contextual modulation, as well as potentially providing insights towards the underlying computational organization of cortical development.

\begin{figure}
    \centering
     \includegraphics[width=1\linewidth]{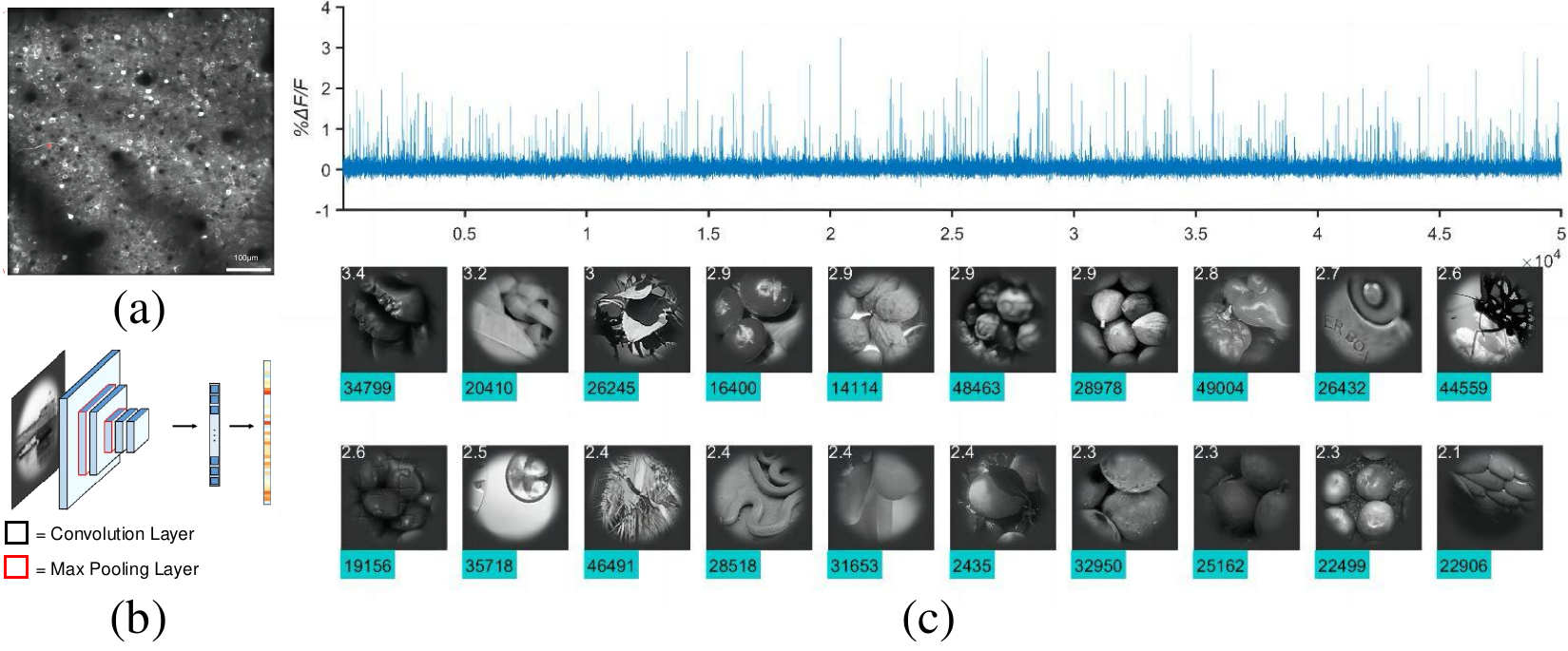}
    \caption{Macaque neuronal response dataset. \textbf{(a)} shows a two-photon image with cells. \textbf{(b)} shows a  feedforward CNN used to model neural response. \textbf{(c)} shows the response of one neuron to $50$k stimuli and the top 20 images that induced the strongest responses. On average, less than 0.5\% of the images induce responses greater than half peak height. Each site contains around 300 neurons.}
    \label{z_introdata}
\end{figure}

\section{Related works}
\paragraph{Modeling neural response prediction} Feedforward deep neural networks have proven effective in modeling and predicting neural responses in early visual brain areas  \citep{10.1167/19.4.29,klindt2018neural,Yamins2016UsingGD,Zhang2018ConvolutionalNN,Cadena201764,annurev:/content/journals/10.1146/annurev-vision-082114-035447}. However, the brain's visual areas contain abundant recurrent connections that are essential for generating neural responses \citep{Felleman1991DistributedHP,https://doi.org/10.1002/cne.23458,Spoerer677237}. Incorporating biologically-inspired simple recurrent circuits, in the form of a Markov network, into convolutional neural networks has been shown to enhance efficiency compared to purely feedforward models, achieving similar performance in image classification and neural prediction tasks \citep{zhang2022recurrentnetworksimproveneural}. In the context of neural prediction, the underlying assumption is that the closer a model can replicate the neural computation mechanisms responsible for a real neuron's response, the more accurate the model's predictive capabilities become \citep{pogoncheff2023explaining,Willeke2023.05.12.540591,li2019learning}.

\paragraph{Self-attention for global dependencies} 

Self-attention mechanisms have recently become a pivotal component in deep learning models, especially in natural language processing and increasingly in computer vision tasks \citep{vaswani2023attention,zhao2020exploring,9522921}. In computer vision, self-attention performs a weighted average operation based on the context of input features, computing attention weights dynamically through a similarity function between pixel pairs \citep{vaswani2023attention,pan2022integration}. This flexibility allows the attention module to adaptively focus on different regions and capture informative features \citep{ramachandran2019standalone}. Self-attention has also been integrated with CNNs to enhance their representational power \citep{pan2022integration,yang2019convolutional,bello2020attention}. By enabling CNNs to consider distant spatial relationships within an image, self-attention improves the network's ability to capture global context. This mechanism overcomes the limitations of traditional CNNs, which primarily concentrate on local features because of their convolutional structure. Taking the complementary properties of convolution and self-attention, the benefits of each paradigm can be extracted by integrating the two and using self-attention to augment convolution modules \citep{dai2021coatnet,yang2019convolutional,pan2022integration,cordonnier2020relationship}. 

\begin{figure}[tbp]
    \centering
    \includegraphics[width=1\linewidth]{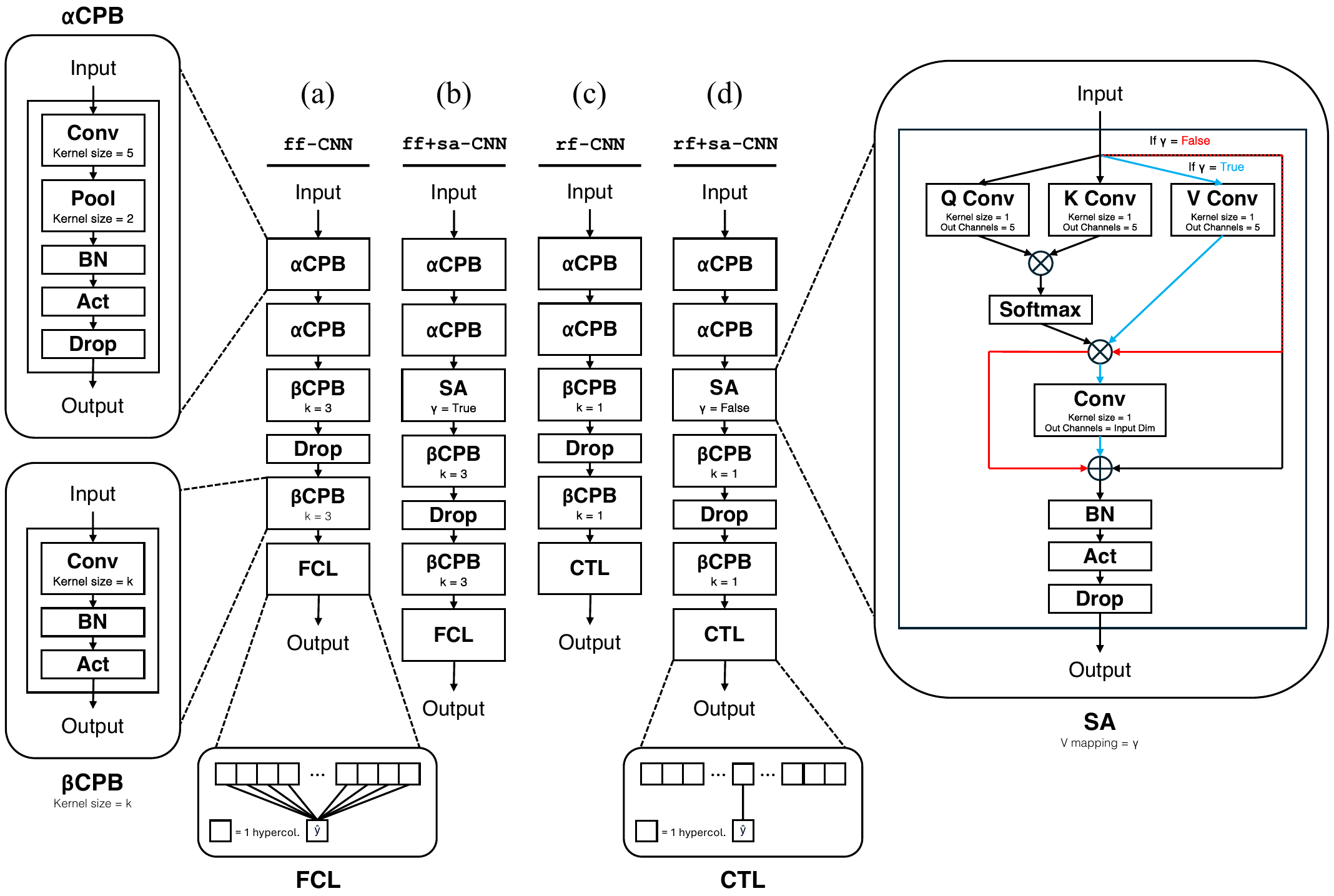}
    \caption{Models explored in this study. Models are constructed from two types of convolutional processing blocks (CPB): $\alpha$CPB and $\beta$CPB. $\alpha$CPB has a fixed convolution kernel size $= 5$ and max pooling kernel size $= 2$. $\beta$CPB takes an input convolution kernel size of $k$, and has no pooling layers. The two final layer readout modes are fully connected (FCL) and center hypercolumn only (CTL). Self-attention (SA) takes as input a boolean $\gamma$ that determines whether the value (V) vector is transformed; if $\gamma = $True then V is mapped, otherwise V is equal to the input. All models with SA utilize single-headed attention. \textbf{(a)} shows the feedforward CNN. \textbf{(b)} shows the feedforward CNN augmented with self-attention. \textbf{(c)} shows the receptive field CNN. \textbf{(d)} shows the receptive field CNN augmented with self-attention.}
    \label{z_arch}
\end{figure}

\section{Approach}

In this study, we developed a set of deep learning models to model V1 neural response to natural images, with the goal of evaluating the potential roles of the self-attention mechanism in neural computation within the visual cortex. We obtained a dataset of neuronal responses measured using two-photon imaging with GCaMP5 from two awake behaving macaque monkeys performing a fixation task, consisting of $302$ neurons from monkey 1 (M1S1) and $299$ neurons from monkey 2 (M2S1), in response to $34$k and $49$k natural images extracted from the ImageNet dataset. The neurons were recorded across five days and tracked anatomically based on landmarks as well as based on their responses to $200$ fingerprint images tested every day. The images were presented in sequence with $500$ ms per image preceded by $1500$ second of blank screen. The $30$k-$50$k images in the training set were presented once, and the $1000$ images in the validation set were tested once with $10$ repeats. Images were $100\times 100$ pixels, with $30\times 30$ pixels for $1$ degree visual angle. The eccentricity of the recording sites were $3$ degrees and $1.79$ degrees, with average receptive field sizes (diameters at half-height) of about $0.75$ and $0.5$ degrees, respectively. We preprocessed the dataset before modeling, and notably downsampled input images to $50\times 50$ pixels, yielding $15\times 15$ pixels per degree visual angle. The receptive fields of neurons from each $1$ mm $\times$ $1$ mm site (the scale/size of a hypercolumn) in macaque monkeys exhibited significant overlap. These fields were mapped using oriented bars or SmoothGrad feature attribution on our deep learning model. The standard deviation of the receptive field centers, all of which are contained within the center hypercolumn of our CNN models, was less than 1 pixel ($1/15 $ degree visual angle). See Appendix \ref{a_exp_setup} for more details of the macaque experimental setup. 

\subsection{Augmenting feedforward CNNs with self-attention}

First, we investigate whether incorporating self-attention into the baseline feedforward CNN model enhances neural response prediction performance. See Appendix \ref{a_comp_model} for comparisons to other established models.

\paragraph{Baseline feedforward model (\texttt{ff-CNN})} See Figure \ref{z_arch}(a) for the architecture. The baseline feedforward model is comprised of two $\alpha$-convolutional processing blocks ($\alpha$CPB) and two $\beta$-convolutional processing blocks ($\beta$CPB), followed by a fully connected readout layer (FCL) . A single \texttt{ff-CNN} model is fitted to each neuron. All models described below are derived from this baseline model. Given a grayscale input image with dimensions $50 \times 50$ pixels, the two $\alpha$CPB layers with a $5\times 5$ kernel encode the input of size $(1\times 50 \times 50)$ into $(c \times 9 \times 9)$ where $c\in \mathbb{N}$ is the number of channels ($c \in \{30,32\}$ in this study). The center hypercolumn of the post-$\alpha$CPB encoding has a centered effective receptive field size of $13\times 13$ pixels. In other words, the center hypercolumn of the latent representation after the $\alpha$CPB layers will have a $13\times 13$ (or $0.8 \times 0.8$ degree visual angle) feedforward receptive field at the center of input $50\times 50$ image. Note that the real neurons' receptive fields are contained inside the receptive field of the center hypercolumn. In the baseline model, the two $\alpha$CPB layers are followed by two $\beta$CPB layers with $3\times 3$ kernels to further expand the effective receptive field of the center-hypercolumn. Finally, \texttt{ff-CNN} has access to the entirety of the input image in the final layer as the readout has full access to all the hypercolumns. Thus, the baseline \texttt{ff-CNN} CNN has two modalities of contextual modulation -- convolutions and a fully connected layer.

\paragraph{Feedforward with self-attention model (\texttt{ff+sa-CNN})} See Figure \ref{z_arch}(b) for the architecture. We augment \texttt{ff-CNN} with a self-attention layer immediately after the last $\alpha$CPB and before the first $\beta$CPB. This placement enables SA to act on an adequately convolved feature representation, but also be further modulated by convolutions before feeding into the final layer. We compare the performance of \texttt{ff+sa-CNN} against that of \texttt{ff-CNN}, controlling the parameter counts to be roughly equal by decreasing the number of channels, which is maintained throughout entire model, from $c=32$ in the baseline CNNs to $c=30$ in the self-attention models to account for the addition of the SA layer. In the context of contextual modulation, \texttt{ff+sa-CNN} intermixes spatial interactions and inter-channel mixing across SA, the posterior $\beta$CPBs, and the FCL.

\subsection{Factorizing the contextual modulation mechanisms}

There are three mechanisms in \texttt{ff+sa-CNN} mediating contextual interactions. We proceed to factorize \texttt{ff+sa-CNN} by removing the contextual modulation contributed by the $\beta$CPBs and the FCL to assess the standalone capability of SA in incorporating surrounding context. This is accomplished by constructing a baseline receptive field model and a model where only SA is mediating horizontal connections. 

\paragraph{Baseline receptive field model (\texttt{rf-CNN})} See Figure \ref{z_arch}(c) for the architecture. We first construct the \texttt{rf-CNN} model, which is devoid of contextual modulation, by subtracting from \texttt{ff-CNN}: the kernel size in the $\beta$CPBs are changed from $3\times 3$ to $1\times 1$ and the fully connected layer is changed to look only at the center hypercolumn (CTL). The $1\times 1$ convolutions perform no spatial expansion before feeding into the CTL. Thus, \texttt{rf-CNN} is making predictions solely based on the center hypercolumn receptive field produced by the $\alpha$CPBs, which covers the center $13 \times 13 $ pixels of the input image. 
 
\paragraph{Receptive field with self-attention model (\texttt{rf+sa-CNN})} See Figure \ref{z_arch}(d) for the architecture. We add self-attention to \texttt{rf-CNN} prior to the $\beta$CPBs to construct \texttt{rf+sa-CNN}. Self-attention is the {\bf only}  mechanism for incorporating surround context in this model. The parameter counts are again controlled by reducing the number of channels from $c=32$ (in \texttt{rf-CNN}) to $c=30$ (see Appendix \ref{a_ctl_chan} for more details). We compare the performance of the two receptive field models, alongside the feedforward models. Note that in \texttt{rf+sa-CNN}, $\gamma $ is False in the SA layer, meaning SA operates exclusively on the horizontal spatial interactions between hypercolumns without any inter-channel mixing. In contrast, $\gamma $ is True in the SA layer of \texttt{ff+sa-CNN}, which allows channel mixing in SA. Channel mixing potentially provides self-attention greater flexibility (see Appendix \ref{a_vmap} and \ref{a_post_sa_chan}).

\subsection{Incremental learning: factorizing the learning process} 
\begin{wrapfigure}{r}{0.5\textwidth}
    \centering
    \hspace{2cm}
    \includegraphics[width=1\linewidth]{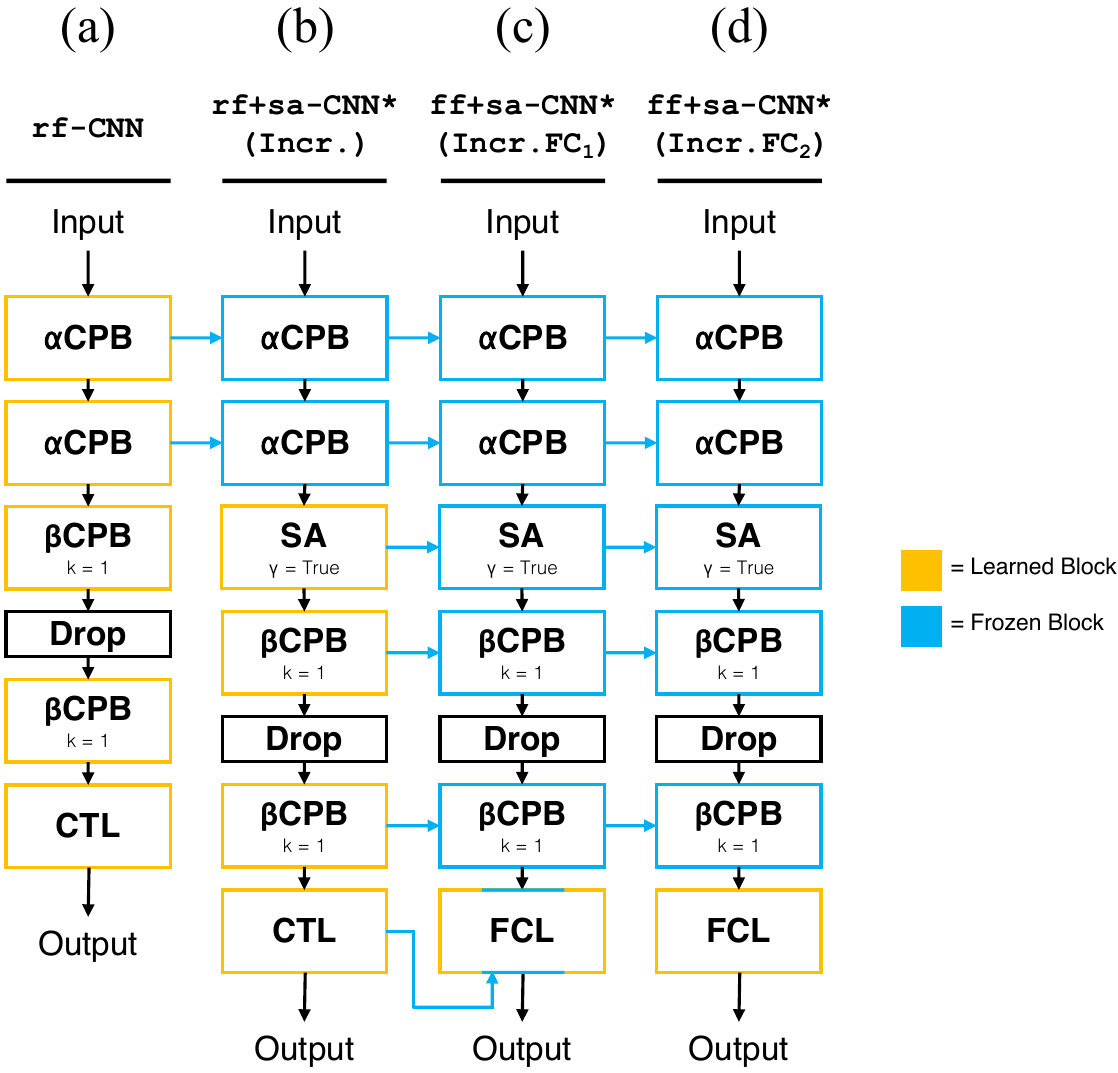} 
    \caption{Incremental learning models. \textbf{(a)} shows the baseline receptive field CNN, equivalent to Figure \ref{z_arch}(c). \textbf{(b)} shows (a) augmented with SA and learned incrementally; the $\alpha$CPBs are taken from (a) and the remaining layers are learned. The $^*$ denotes slight modification from \texttt{rf+sa-CNN}, Figure \ref{z_arch}(d), namely $\gamma$ is changed to True. \textbf{(c), (d)} show the result of replacing the CTL in (b) with a FCL, and learned incrementally; (c) freezes only the center hypercolumn in the FCL (\texttt{FC$_1$}) whereas (d) allows the FCL to learn freely (\texttt{FC$_2$}). (c) and (d) have all other layers taken from (b). The $^*$ denotes slight modification from \texttt{ff+sa-CNN}, Figure \ref{z_arch}(b), namely $k$ in $\beta$CPB is changed to $k=1$. \texttt{(Simul.)} models are equivalent in architecture, except all blocks are learned.}
    \label{z_incmodel}
\end{wrapfigure}

Discussed in Section \ref{incsec}, comparing the four models above reveals that contextual modulation introduced in \texttt{rf+sa-CNN} via SA did not produce better performance relative to \texttt{rf+CNN}, despite \texttt{ff+sa-CNN} having clear performance improvements over \texttt{ff-CNN}. This is not due to the difference in channel mixing (the $\gamma$ parameter in SA). We hypothesize that bottle-necking all the gradient signals solely through the center hypercolumn during backpropagation makes it difficult for the network to properly learn the $\alpha$CPB layers and the SA layer simultaneously. Thus, we investigate an incremental learning paradigm where we allow the receptive fields of the $\alpha$CPBs to be learned before incorporating any context mechanisms. We then incrementally add and learn a self-attention layer followed by a fully connected readout layer.

The following progression of models, \texttt{rf-CNN}, \texttt{rf+sa-CNN$^*$}, and \texttt{ff+sa-CNN$^*$} (as shown in Figure \ref{z_incmodel}), incrementally expands the capacity of contextual modulation. An important distinction between incremental models and models shown in Figure \ref{z_arch}, marked by $^*$, is a $1\times 1$ kernel in the $\beta$CPB, which maintains channel mixing but removes further spatial integration through convolution. \texttt{rf-CNN} (shown in Figure \ref{z_incmodel}(a) or Figure \ref{z_arch}(c)) has information only from the center receptive field. \texttt{rf+sa-CNN$^*$} (shown in Figure \ref{z_incmodel}(b)) uses only the self-attention mechanism for contextual modulation. \texttt{ff+sa-CNN$^*$} (shown in Figure \ref{z_incmodel}(c)-(d)) has the same surround-center modulation as \texttt{rf+sa-CNN$^*$} from self-attention, but allows spatial integration of the global context by changing the CTL to FCL at the end.  As horizontal connections in the visual cortex are known to mature after the development of the receptive fields, we designed an incremental learning setup where \texttt{rf-CNN} first learns the receptive fields, then \texttt{rf+sa-CNN$^*$(Incr.)} learns a self-attention layer only after \texttt{rf-CNN} has already learned the $\alpha$CPB receptive fields. Finally, \texttt{ff+sa-CNN$^*$(Incr.FC$_1$)} and \texttt{ff+sa-CNN$^*$(Incr.FC$_2$)} inherit the receptive fields and self-attention structures of \texttt{rf+sa-CNN$^*$}, but differ in the change to a FCL readout. Models labelled \texttt{(Incr.)} are learned incrementally as such, and models labelled \texttt{(Simul.)} are traditionally trained simultaneously.

\subsection{Hyperparameter selection and Model training} 
Rather than splitting the evaluation set for hyperparameter selection, we partitioned our population of neurons to select hyperparameters (training and architectural). We fine-tuned, by experimenting with batch size, learning rate, epochs, number of layers, number of channels per layer, etc., model hyperparameters on a subset of 50 neurons using a relatively coarse grid search. We list key training hyperparameters here: (1) batch size = $50$, (2) learning rate = $0.001$, (3) optimizer = Adam, (4) loss = MSE, (5) epochs = $50$. Training and computations were performed on an in-house computing cluster with GPU (NVIDIA V100 or similar) nodes. Training hyperparameters were held constant across all models. Architectural hyperparameters were held constant across layers shared between models. We do not optimize hyperparameters for models other than the baseline \texttt{ff-CNN}. 

The primary objective of this project is to demonstrate that self-attention can enhance neural response prediction relative to the baseline feedforward CNN, despite hyperparameters being optimized only for the baseline model. Since we show that \texttt{ff+sa-CNN} improves upon \texttt{ff-CNN} in both evaluation metrics (see Section \ref{corrsec}), further hyperparameter optimization is unnecessary for our objective. Instead, we are interested in understanding the reason behind this improvement. The other models tested in this study are architectural subsets of the \texttt{ff+sa-CNN}, designed to dissect their contributions to its success. We do not anticipate any derivative models outperforming \texttt{ff+sa-CNN}, justifying holding hyperparameters constant across models for fair comparison.

\subsection{Model evaluation metrics}
To quantify performance, models were evaluated on two criteria, Pearson correlation and peak tuning index. Pearson correlation represents the overall tuning similarity between a model's predicted responses and the real neuron's recorded responses. The peak tuning index is used to quantify how well a model can predict and match in magnitude the strongest responses recorded by the real neuron. This lets us evaluate how well a model can discriminate between, as well as model the response magnitude of, images that are strongly excitatory and images that incite a weak response.

\begin{figure}[t]
    \centering
    \includegraphics[width=0.32\linewidth]{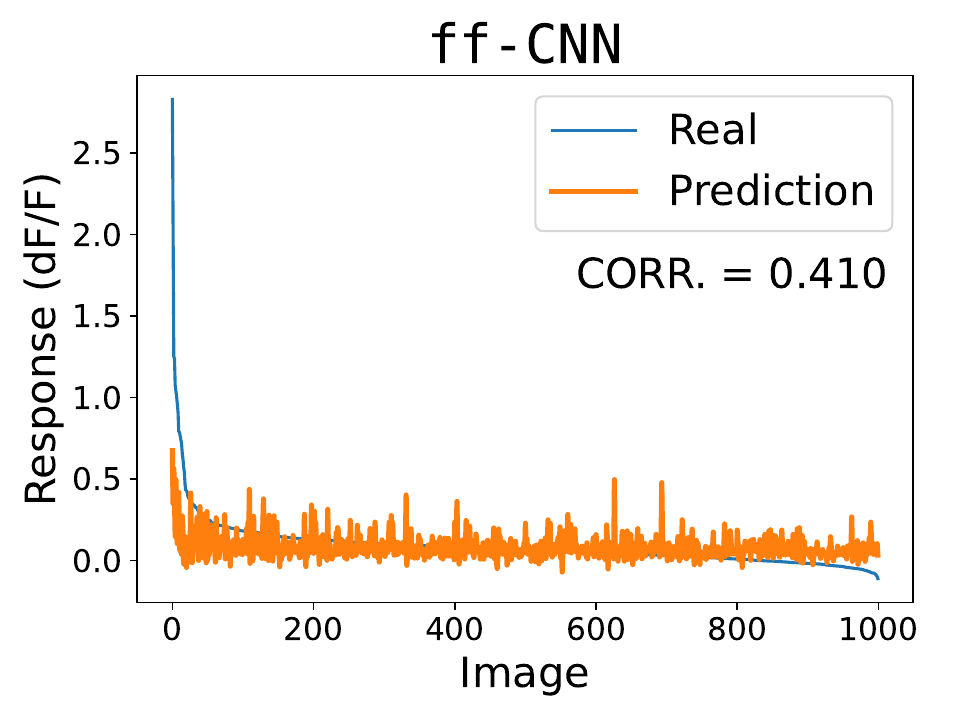}
    \includegraphics[width=0.32\linewidth]{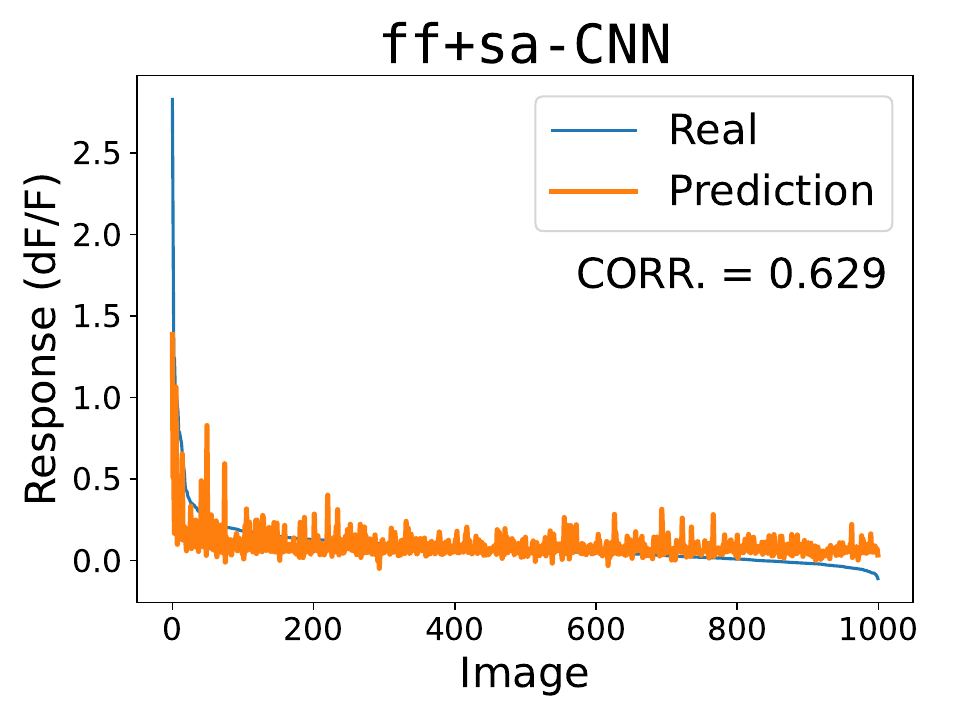}
    \includegraphics[width=0.32\linewidth]{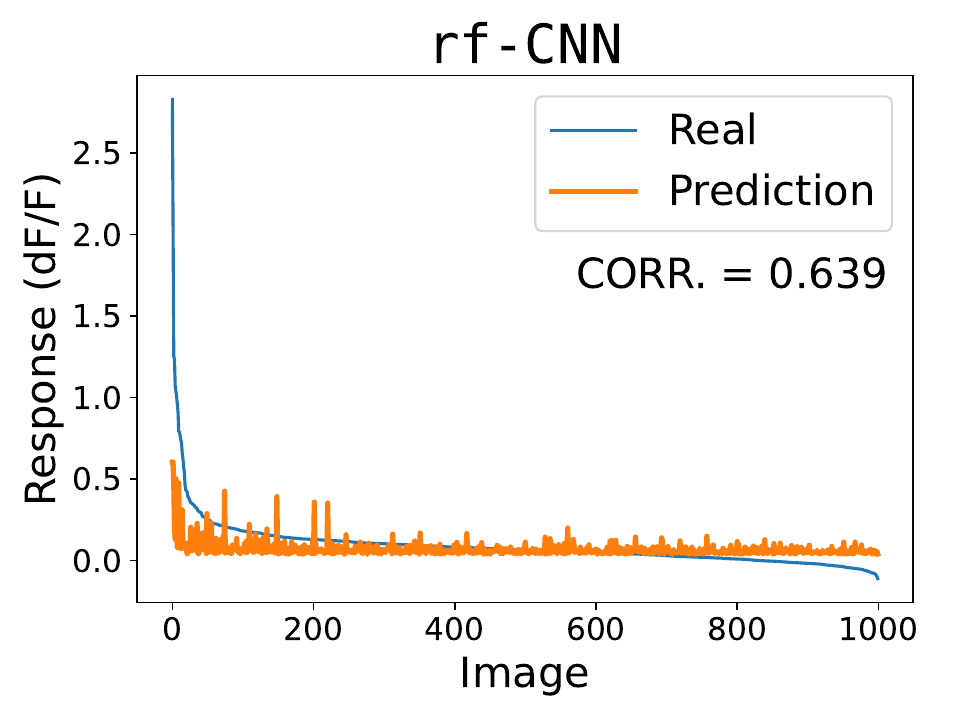}
    \caption{Neuronal tuning curves of \texttt{ff-CNN}, \texttt{ff+sa-cnn}, and \texttt{rf-CNN}. Pearson correlation does not reflect peak tuning. Despite \texttt{rf-CNN} having the better correlation, it is clear that \texttt{ff+sa-CNN} is able to capture the peak significantly better, at the cost of a noisier overall tuning. Example shown is M1S1 neuron 238. See Appendix \ref{a_pop_tc} for population averages.}
    \label{z_tc}
\end{figure}

\paragraph{Pearson correlation:} The Pearson correlation (CORR.) is taken between neuron responses and model-predicted responses. Pearson correlation is a standard measure for evaluating neural response prediction. Other established measures, including FEV, $r^2_{er}$, and CC$^2_{norm}$ \citep{pmlr-v220-willeke23a,Cadena201764, 10.1371/journal.pcbi.1009212, zhang2022recurrentnetworksimproveneural}, were used as well and yielded similar results to Pearson correlation (see Appendix \ref{a_comp_model} and \ref{a_comp_metric}).

In this paper, we explored a set of measures to assess the peak of a neuronal tuning curve. In our macaque V1 dataset, we found that neurons exhibit sharp stimulus selectivity, consistent with findings from \citet{tangelife, tangcb}, reinforcing the diversity and complexity of V1 neurons (see Appendix \ref{a_feat}). We found that Pearson correlation and other standard metrics (see Appendix \ref{a_comp_metric}) are successful in measuring a model's fit to the overall tuning curve, but often fail to represent the peak tuning preference of neurons, of which is a key aspect of a neuronal function. For example, Figure \ref{z_tc} demonstrates that while \texttt{rf+CNN} achieves a higher correlation in approximating the overall tuning curve compared to \texttt{ff+sa-CNN}, the latter outperforms in modeling the neuron's tuning curve peak. To address this issue, we developed two new metrics to better assess a model's ability to capture the peak tuning of neurons.

\paragraph{Peak tuning index: }The peak tuning index (PT) is a membership metric of the strongest predictions above a threshold determined by the top $1\%$ of real responses. PT can be roughly interpreted as the percentage of the peak that a model captures, under a magnitude prior. The index is calculated as: 
$$\text{PT} = \frac{\texttt{\# of top 1\% predictions} \geq \texttt{min(top 1\% real responses)}}{\texttt{\# of responses in the top 1\%}} \times \texttt{100\%}$$
PT is  divided into PT$_J$ and PT$_S$, based on how \texttt{\# of top 1\% predictions} is defined. PT$_J$ is when predictions are jointly rank ordered with respect to the real responses. PT$_S$ is when predictions are separately rank ordered independently of the real responses. PT$_J$ is a stricter measure. Note that because we train with MSE loss, models are incentivized to minimize the absolute difference between predictions and real responses, rather than match the curvature of the tuning curve. This minimizes the risk of PT being misrepresentative due to lateral shifts in the tuning curve.

\section{Results}

\subsection{Self-attention improves neural response prediction}
\label{corrsec}
\begin{table}[htbp]
  \centering
  \scriptsize  
    \caption{Average Pearson correlation and peak tuning metrics for models trained on M1S1 and M2S1. Correlation SEM $=0.009$ was consistent across models and monkeys. Despite $\texttt{rf-CNN}$ unexpectedly outperforming $\texttt{rf+sa-CNN}$, the difference is recovered when $\texttt{rf+sa-CNN}$ is trained incrementally (see Section \ref{incsec}).
    } 
  \label{corr}
  \begin{tabular}{lcccccccc}
    \toprule
    & \multicolumn{4}{c}{M1S1} & \multicolumn{4}{c}{M2S1} \\
    \cmidrule(lr){2-5} \cmidrule(lr){6-9}
    Model & CORR.      & $\Delta$ \texttt{ff-CNN} &  PT$_J$     & PT$_S$ & CORR.  & $\Delta$ \texttt{ff-CNN} & PT$_J$ & PT$_S$\\
    \midrule
    \texttt{ff-CNN}     & $0.393$   & $0.0\%$  & $3.3 \pm 0.5$ & $5.6\pm0.9$  & $0.477$ & $0.0\%$   & $8.6\pm0.9$ & $16.2\pm 1.6$ \\
    \texttt{ff+sa-CNN} & $\mathbf{0.416}$   & $+6.6\%$ & $\mathbf{5.6} \pm 0.6$ & $\mathbf{10.5}\pm1.1$ & $\mathbf{0.491}$ & $+3.3\%$  & $\mathbf{11.5} \pm 0.9$ & $\mathbf{23.5} \pm 1.8$\\
    \midrule
    
    \texttt{rf-CNN}    & $\mathbf{0.420}$ & $+8.6\%$ & $\mathbf{1.1} \pm 0.3$ & $\mathbf{1.8}\pm0.5$   & $\mathbf{0.496}$ & $+4.3\%$  & $\mathbf{4.4} \pm 0.6$ & $\mathbf{6.6} \pm 1.0$\\
    \texttt{rf+sa-CNN} & $0.414$   & $+7.2\%$ & $0.7 \pm 0.2$ & $1.0 \pm 0.3$  &  $0.486$ & $+2.4\%$  & $3.4 \pm 0.5$ & $5.1 \pm 0.8$\\
    

    \bottomrule 
  \end{tabular}
\end{table} 

We compared the performance of the \texttt{ff+sa-CNN} model to the parameter-matched baseline \texttt{ff-CNN} model and found that incorporating self-attention significantly improved correlation and both peak tuning metrics (see first two rows of Table \ref{corr}). This indicates that self-attention enhances modeling of both the overall tuning and peak tuning aspects of the neurons, with consistent results across both monkeys.

It is important to note that the \texttt{ff-CNN} center hypercolumn at the readout layer has a receptive field much larger than the real neuron's receptive field, due to successive convolutions in the $\alpha$CPBs and $\beta$CPBs. Additionally, the fully connected readout layer also incorporates long-range spatial dependencies. Thus, the self-attention layer in \texttt{ff+sa-CNN} acts as an additional mechanism for modeling horizontal connections, and provides additional performance benefits. 

To better understand the role of self-attention in contextual modulation, we constructed a baseline receptive field model, \texttt{rf-CNN}, that is devoid of all contextual modulation mechanisms. \texttt{rf-CNN}'s CTL readout only uses the center-hypercolumn of the convolved feature space to make predictions. Note that the feedforward receptive field of the center hypercolumn after the two $\alpha$CPBs is a centered $13\times 13$ pixel portion of the input image. Moreover, between the $\alpha$CPB layers and the CTL layer, the size of the center hypercolumn receptive field does not expand due the the $1\times1$ kernel in the $\beta$CPBs. This means that predictions are being made solely based on the center $13\times 13$ pixels, which corresponds roughly to the real neuron's receptive field.

Surprisingly, we found that \texttt{rf-CNN} achieved the highest correlation, indicating that models focusing primarily on the classical receptive field offer the best fit to the overall tuning curve (see Table \ref{corr}). However, correlation fails to reflect the model's shortcomings in fitting the peak of the tuning curve. Table \ref{corr} also reveals that while \texttt{rf-CNN} has the highest Pearson correlation, it performs worse at capturing the tuning peak compared to \texttt{ff+sa-CNN} and \texttt{ff-CNN}. Thus, we conclude that contextual modulation plays a crucial role in peak tuning, and that the three mechanisms in \texttt{ff+sa-CNN} for integrating surround information are complementary.

\subsection{Dissecting Contextual Modulation Mechanisms via Incremental Learning} 
\label{incsec}

We explore the relative contributions of the different contextual modulation mechanisms. Specifically, we first ask: Is self-attention alone sufficient to model contextual modulation? To answer this question, we added SA to \texttt{rf-CNN} to produce  \texttt{rf+sa-CNN}. We found that without posterior $\beta$CPBs and a FCL, the self-attention in \texttt{rf+sa-CNN} is not useful. In fact, the performance is worse than \texttt{rf-CNN} in overall correlation and both peak tuning metrics (see Table \ref{corr}). This result is somewhat unexpected, as the addition of self-attention in \texttt{ff+SA-CNN} does improve upon \texttt{ff-CNN}. It is possible that the spatial integration mechanisms in the $\beta$CPBs, along with the FCL, are necessary to provide sufficient pathways for backpropagating the gradients during learning, so that both the $\alpha$CPB receptive fields and self-attention kernels can be properly learned.

To test this hypothesis, we explored whether an incremental learning approach, where different network components are learned sequentially, could yield a 
model that performs on par with \texttt{ff+sa-CNN}. We show that learning the feedforward kernels in the $\alpha$CPBs first, followed by learning the self-attention layer, and finally the fully connected layer, can nearly match the performance of \texttt{ff+sa-CNN}. This indicates that although the spatial integration by $\beta$CPBs contributes to performance, the self-attention layer plays a more critical role in capturing horizontal interactions crucial for modeling peak tuning.

\begin{table}[htbp]
  \centering
    \caption{Average Pearson correlation for models incrementally and simultaneously trained on M1S1 and M2S1. Correlation SEM $=0.009$ was consistent across models and monkeys.}
  \label{inccorr}
  \scriptsize %
  \begin{tabular}{lcccc}
    \toprule
     & \multicolumn{2}{c}{M1S1} & \multicolumn{2}{c}{M2S1} \\
    \cmidrule(lr){2-4}\cmidrule(lr){4-5}
     Model (Training Method) & CORR. & $\Delta$ \texttt{rf-CNN}  & CORR. & $\Delta$ \texttt{rf-CNN} \\
    \midrule
    \texttt{rf-CNN(Simul.)}     & $\mathbf{0.420}$   & $0.0\%$ & $\mathbf{0.496}$ & $0.0\%$ \\
    \midrule
    \texttt{rf+sa-CNN$^*$(Simul.)}  & $0.409$   & $-2.6\%$& $0.480$ & $-3.2\%$\\
    \texttt{rf+sa-CNN$^*$(Incr.)}     & $\mathbf{0.421}$   & $+0.6\%$& $\mathbf{0.493}$ & $-0.3\%$ \\
    \midrule
    \texttt{ff+sa-CNN$^*$(Simul.)} & $0.416$   & $-0.8\%$ & $0.490$ &$-0.7\%$ \\
    \texttt{ff+sa-CNN$^*$(Incr.FC$_1$)} & $\mathbf{0.430}$   & $+3.0\%$ & $\mathbf{0.494}$ &$-0.1\%$ \\
    \texttt{ff+sa-CNN$^*$(Incr.FC$_2$)} & $0.414$   & $-1.3\%$ & $0.488$ &$-1.1\%$  \\
    \bottomrule
  \end{tabular}
\end{table}

\begin{figure}[tbp]
    \centering
    \includegraphics[width=0.47\linewidth]{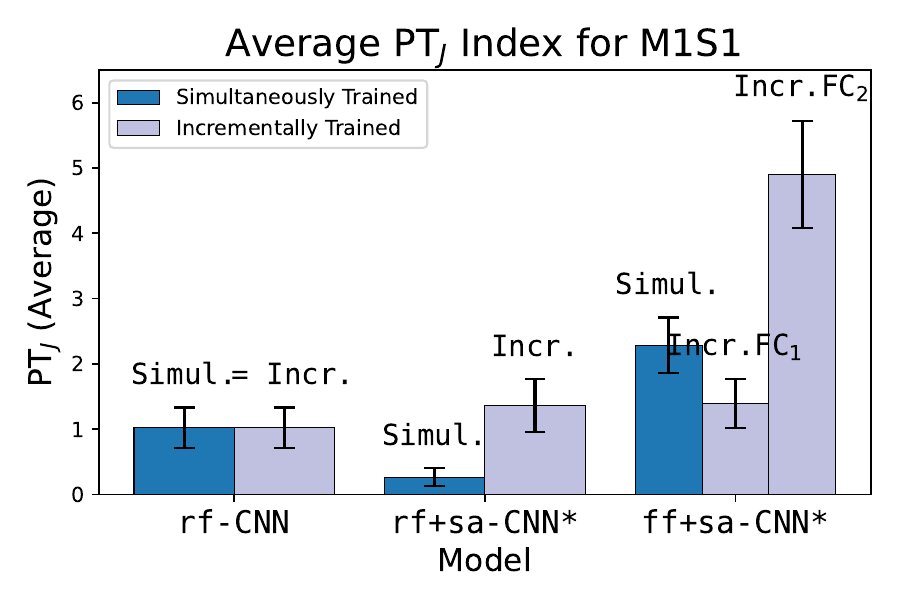}
    \includegraphics[width=0.47\linewidth]{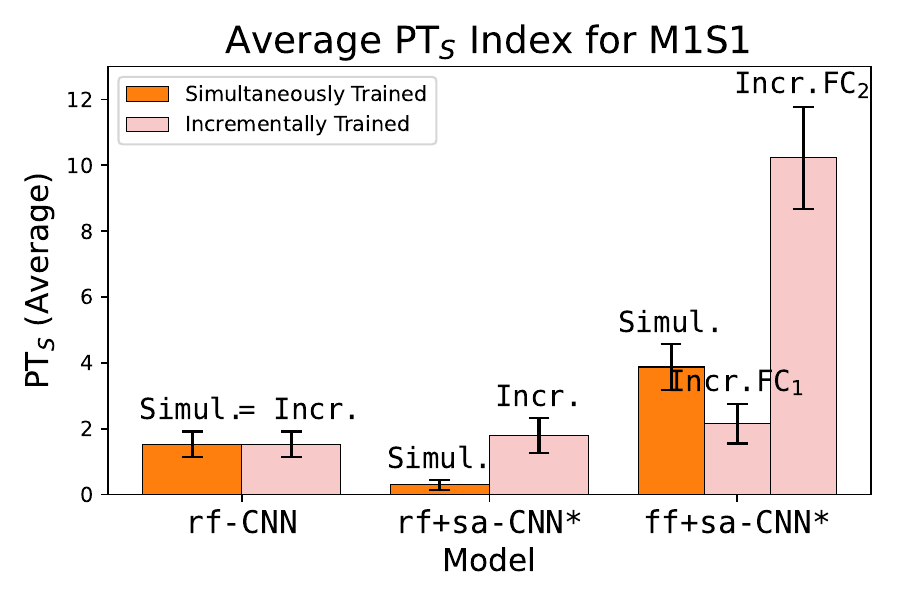}
    \\
    \includegraphics[width=0.47\linewidth]{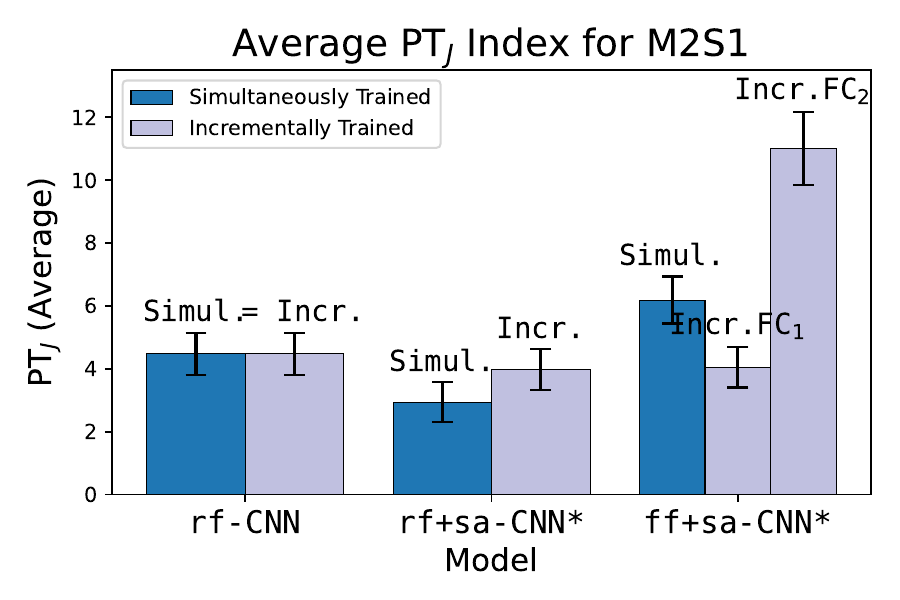}
    \includegraphics[width=0.47\linewidth]{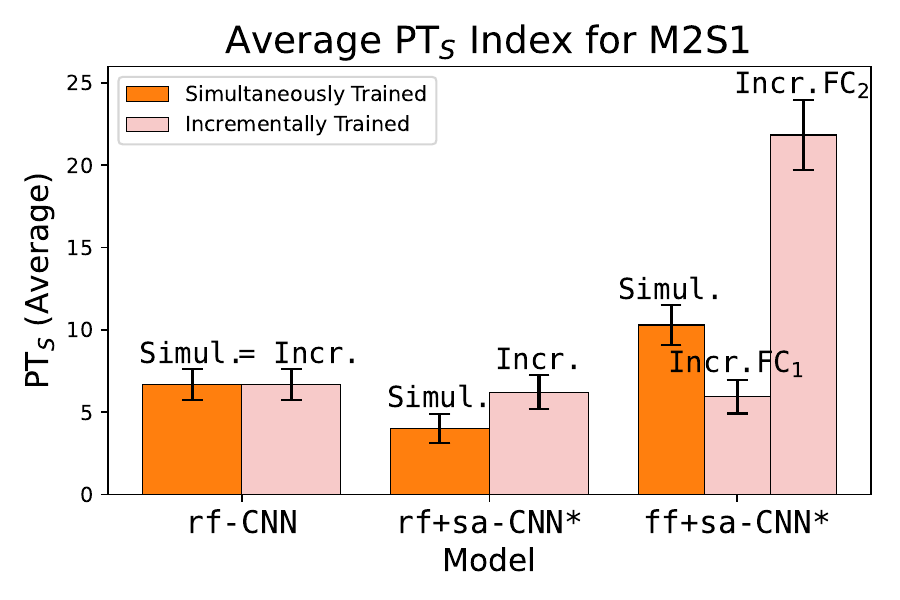}
    \caption{Average peak tuning indices for incrementally and simultaneously trained models. \textbf{Top row:} bar charts for M1S1. \textbf{Bot row:} bar charts for M2S1. \textbf{Left col:} average PT$_J$ values. \textbf{Right col:} average PT$_s$ values. Error bars are SEM.}
    \label{incpt}
\end{figure}

Incremental learning offers a valuable approach for accurately assessing the potential of each spatial integration mechanism in modeling contextual modulation and evaluating their interdependence in generating an effective model. Table \ref{inccorr} highlights several incremental models we tested and their relative improvements. Learning the receptive field first, followed by learning  self-attention in \texttt{rf+SA-CNN$^*$(Incr.)}, outperforms learning both simultaneously in \texttt{rf+SA-CNN$^*$(Simul.)} (see second and third rows of Table \ref{inccorr}, and middle pairs in Figure \ref{incpt}). This supports our hypothesis that jointly learning the $\alpha$CPB receptive fields and self-attention may overwhelm the system when gradients are constrained to propagate through the center hypercolumn alone. Incremental learning can improve \texttt{rf+sa-CNN$^*$} to match the peak prediction performance of \texttt{rf-CNN}, but not beyond. This suggests that, when used with a CTL readout, self-attention alone is insufficient to fully capture peak tuning.

Table \ref{inccorr} and Figure \ref{incpt} demonstrate that removing the CTL restriction--i.e., allowing the readout to access information from the hypercolumns in the final convolutional layer via FCL, as in \texttt{ff+sa-CNN$^*$}--enables the network to nearly match the performance of the \texttt{ff+sa-CNN}. \texttt{ff+sa-CNN$^*$} is named as such because it closely resembles \texttt{ff+sa-CNN}, with the only difference being the use of a $1\times 1$ kernel instead of a $3\times 3$ kernel in the $\beta$CPBs. These findings suggest that posterior convolution is not required for spatial integration after the self-attention layer when the model is trained incrementally.

A reoccurring observation is that focusing on receptive field information tends to improve overall correlation, while emphasizing contextual information enables the network to better model peak tuning. This pattern is evident when comparing \texttt{rf-CNN} with \texttt{ff+sa+CNN} in Table \ref{corr}. A similar contrast exists between \texttt{ff+sa-CNN$^*$(Incr.FC$_1$)} and \texttt{ff+sa-CNN$^*$(Incr.FC$_2$)}. In \texttt{ff+sa-CNN$^*$(Incr.FC$_1$)}, the model inherits the center hypercolumn weights and then learns the surrounding hypercolumn contributions through the FCL. In contrast, \texttt{ff+sa-CNN$^*$(Incr.FC$_2$)} learns the weights of all hypercolumns in the FCL simultaneously. While the former excels in correlation, the latter performs better in peak tuning. This suggests that the receptive field is most important towards overall tuning, whereas surround-center interactions are key to capturing peak tuning.

The \texttt{ff+sa-CNN$^*$} models saw an improvement in PT$_J$ and PT$_S$ over the \texttt{rf+sa-CNN$^*$} models. This suggests that either the fully connected layer (FCL) plays a critical role in predicting peak responses, or that constraining the readout to the center hypercolumn (CTL) in \texttt{rf+sa-CNN$^*$} limits error propagation to the self-attention block during training. As a result, self-attention may be inadequately learned in these cases, impairing the model's ability to effectively implement contextual modulation.

\subsection{Incremental learning emphasizes the contribution of the classical receptive field}

\begin{figure}[tbp]
    \centering
    \includegraphics[width=0.329\linewidth]{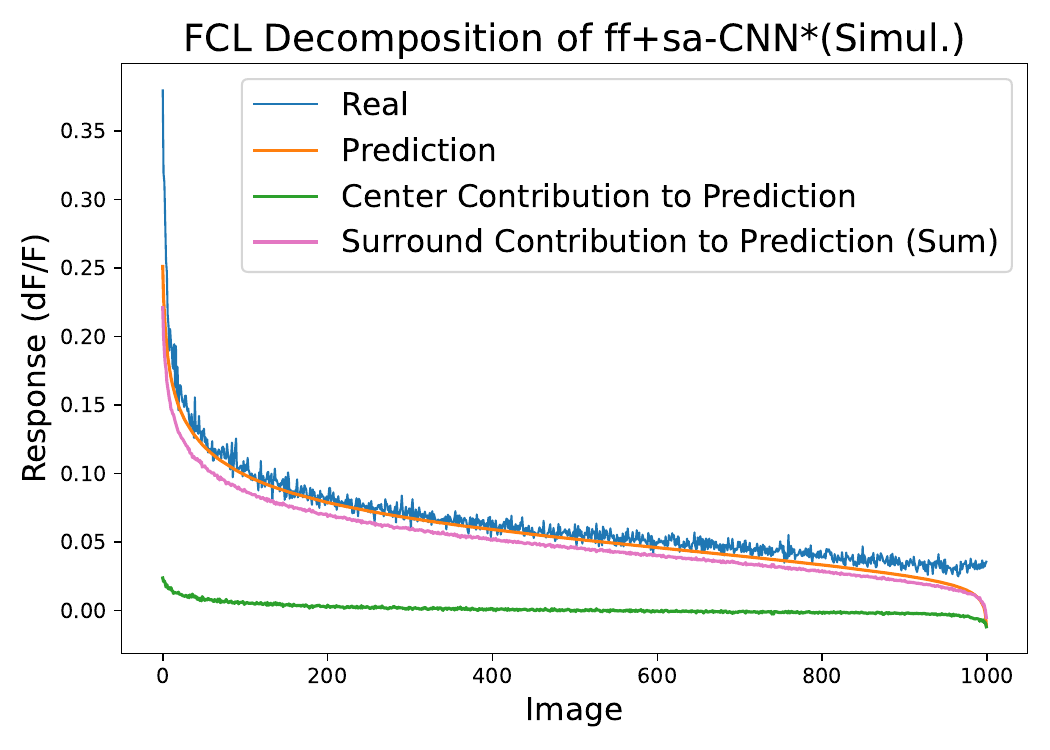}
    \includegraphics[width=0.329\linewidth]{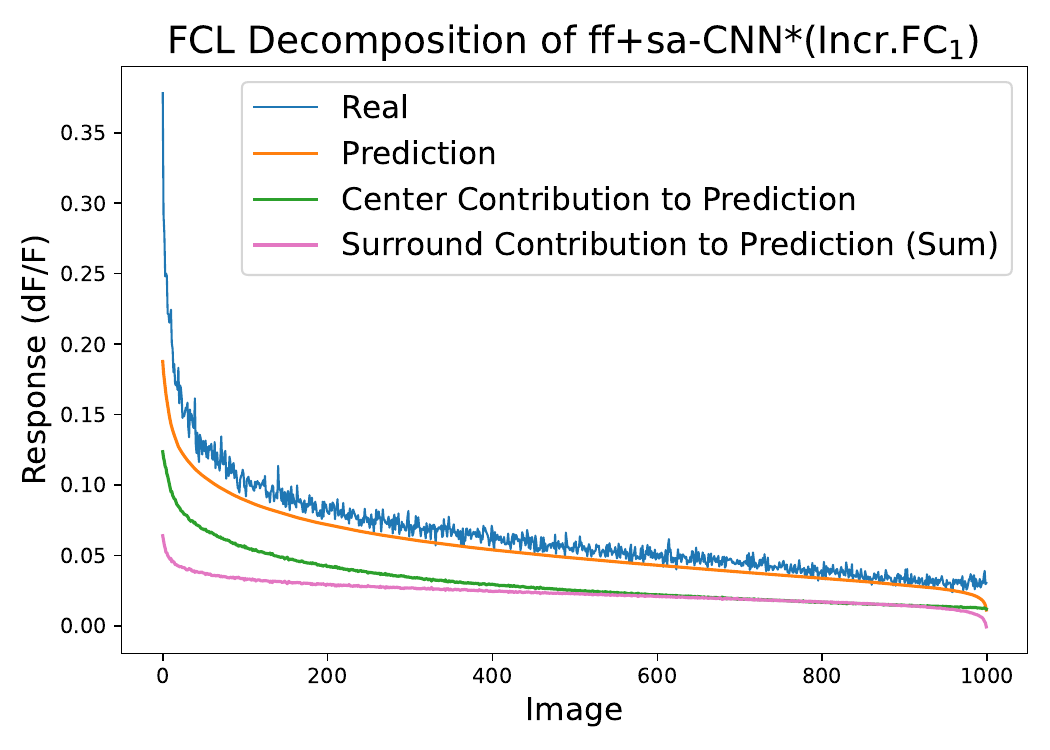}
    \includegraphics[width=0.329\linewidth]{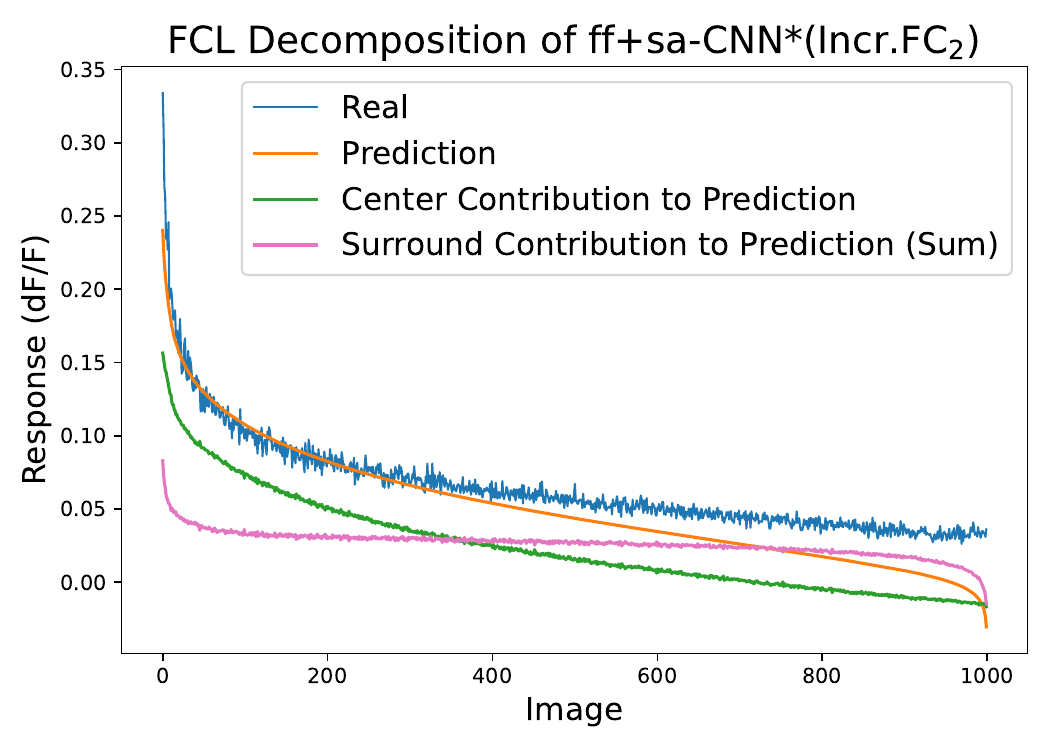}
    \caption{Average FCL decomposition of \texttt{ff+sa-CNN$^*$} when trained differently. The  center contribution (green) and the total surround contribution (pink) sum to the prediction tuning curve (orange). Plots are rank ordered with respect to predicted responses. Averages are calculated by plotting rank ordered decomposed tuning curve for each neuron, then averaging across each image. Individual contributions from hypercolumns can be found in Appendix \ref{a_fcl_decomp}.}
    \label{fcld}
\end{figure}

A well-established neurophysiological principle is that stimuli within the classical receptive field of V1 neurons are the primary driver of neural responses, while the contextual surround modulates them. We found that when the entire network is trained simultaneously, as in \texttt{ff+sa-CNN$^*$(Simul.)}, performance is weaker and the network does not follow this principle. However, with incremental learning, the center hypercolumn develops into the dominant contributor, as observed in \texttt{ff+sa-CNN$^*$(Incr.FC$_1$)} and \texttt{ff+sa-CNN$^*$(Incr.FC$_2$)}. Figure \ref{fcld} illustrates the sum of connection weights from the readout, showing that in \texttt{ff+sa-CNN$^*$(Simul.)}, weights are evenly distributed, whereas in the incrementally trained models, the center hypercolumn, corresponding to the classical receptive field, is emphasized. Incremental learning fosters this center-surround division and modulation, yielding interpretable performance benefits.

\section{Discussion}

CNNs are widely used and effective models for visual cortical neurons, and they inherently include two mechanisms for contextual modulation: successive convolutions and a fully connected layer, which allow the global image context to be accessible to the readout. In this paper, we demonstrated that augmenting CNN models of cortical neurons with self-attention enhances predictions of both the overall tuning curve and the tuning peak. Self-attention, resembling three-way interactions in probabilistic graphical models, facilitates flexible surround-center modulation via contextual variables \citep{article,fei2022attention}. This provides additional flexibility and complementary benefits to the CNN's inherent context mechanisms. While large-scale transformer models with multiple attention heads have achieved state-of-the-art performance in modeling mouse V1 neurons by capturing long-range dependencies \citep{li2023v1tlargescalemousev1}, our work explicitly explores the role of self-attention in CNNs for modeling horizontal circuits, highlighting the dependencies and complementary interactions between different mechanisms of contextual modulation.

Several key findings emerged from this work that advance our understanding of cortical computation and neural codes. First, we found that focusing on receptive field information, as in \texttt{rf-CNN}, yields the highest Pearson correlation, alongisde other standard measures (see Appendix \ref{a_comp_metric}), for overall neuronal tuning curves (see Table \ref{corr}). This suggests that the classical receptive field is the primary driver behind a neuron's overall response. Our incremental learning experiments further supports the advantage of concentrating on information within the classical receptive field in the center hypercolumn (see Figure \ref{fcld}). Second, we demonstrated that contextual modulation is crucial for a strong and robust peak tuning, with self-attention playing a pivotal role. A trade-off, however, exists between the receptive field and surround modulation: RF-centric models fit overall tuning curves more accurately, while increased contextual surround modulation enhances peak tuning, though often at the expense of overall tuning correlation. Incremental learning, which fosters a strong receptive field bias, may help even out this trade-off. This is consistent with neurophysiological evidence supporting a dominant classical receptive field and weaker surround modulation, with recurrent connections potentially being fine-tuned after receptive field development.

A recent CNN-based model of mouse V1 neurons revealed that the most excitable images often involve stimulus features outside the receptive fields, consistent with the concept of pattern completion \citep{Fu2023.03.13.532473}. Similarly, we found that models capable of capturing peak tuning display interpretable contextual modulation, such as association fields and pattern completion, within the self-attention module (see Appendix \ref{a_att}). Additionally, incorporating a self-attention layer improved models' data efficiency (see Appendix \ref{a_de}). Further theoretical and experimental investigations are needed to characterize and evaluate the interactions facilitated by self-attention, in order to uncover how these mechanisms may be implemented by biological circuits.

\section{Ethics Statement}
All procedures involving animals for generating the data of this paper were in accordance with the Guide for the Institutional Animal Care and Use Committee (IACUC) of Peking University.

\section{Acknowledgments}
This work was supported by NSF CISE RI 1816568 and NIH R01 EY030226-01A1 awarded to Tai Sing Lee. Isaac Lin is supported by the NSF REU supplement of RI 1816568. Imaging data was produced by Tianye Wang and Shiming Tang with the support of STI2030-Major Projects 2022ZD0204600, National Natural Science Foundation of China U1909205, and funds from the Peking-Tsinghua Center for Life Sciences to ST.

\newpage
\bibliography{iclr2025_conference}
\bibliographystyle{iclr2025_conference}

\newpage
\appendix
\section{Appendix}
\subsection{Additional Details on Macaque Experimental Setup}
\label{a_exp_setup}
We collected data using a nearly identical experimental protocol as detailed in \citep{tangelife,tangcb}, except that our dataset was considerably larger, including up to 34K and 49K stimulus-response pairs for the two monkeys respectively. 

During each fixation task, a blank screen was presented for 1500 ms after the monkey established fixation, followed by the presentation of a visual stimulus for 500 ms. 34,000 stimuli were tested in monkey 1 and 49,000 stimuli were tested in monkey 2 over a data collection period of 5 days.Neurons were registered anatomically and also by testing a 200 stimuli finger-print each day. To quantify the neural response, a differential image of GCamp5s calcium signals between the stimulus period and blank period was computed for each trial. The dF/F was then calculated based on a 200 ms to 600 ms window after stimulus onset. We note that GCamp5s is slow, but the signal has been found to be correlated with firing rate \citep{LI20171049}. The training set was collected with 1 repeat for each stimulus. The validation set consists of 1000 stimuli, each with 10 repeats.

\subsection{V1 Preferred Features}
\label{a_feat}
\begin{figure} [htbp]
    \centering
    \includegraphics[width=1\linewidth]{shang_poster_fig.pdf}
    \caption{V1 neuron exhibit diversity and complexity in preferred features.
Individual CNNs were used to model neurons' responses to a large set of natural images. \textbf{Mid row:} visualizations of the optimal stimuli for \texttt{ff-CNN} models for 28 neurons. Neurons are clustered into 7 equally sized classes representing preferences for curves, rings (eyes), textures, grating, bars, corners, and other more complex higher order features. \textbf{Top row:} top 5 natural images in the validation set that elicited the largest response from real neurons in each class. \textbf{Bot row:} top 5 artificial stimuli that elicited the largest response from \texttt{ff-CNN} models of neurons in each class. The artificial stimuli preferences are consistent with the optimal stimuli seen in the middle row.}
    \label{v1_feat}

\end{figure}

A driving motivation of our study was to identify the natural image features that V1 neurons prefer, which include corners, curvatures, junctions, rings and other higher order features, rather than the traditional orientation and frequency tunings. While traditional artificial stimuli produce a tuning curve, these stimuli rarely represent the neuron’s most preferred stimulus. The studies referenced above show that the neurons’ true preferences are not necessarily a specific orientation or spatial frequency, and may not be well modeled by Gabor functions and traditional linear/nonlinear models. Rather, the preferred features, which are encoded by the peak responses of neurons as visualized in Figure \ref{v1_feat}, exhibited diversity and complexity in the tuned features. The objective of this study is not to prove or argue for the diversity and complexity of V1 neural codes, but to explore the mechanisms contributing to them. A key finding of this paper is that the surround contextual mechanism, implemented by a non-local network (i.e. self-attention), positively contributes to the generation of the response peak, and is related to the encoding of preferred higher-order features in neurons. As such,this study aims to model the peak response of neurons and develop a novel metric to evaluate the model's accuracy in capturing the peak of each neuron's "natural image tuning curve."


\newpage
\subsection{Comparisons to Other Established Models}
\label{a_comp_model}
\begin{figure} [H]
    \centering
    \includegraphics[width=1\linewidth]{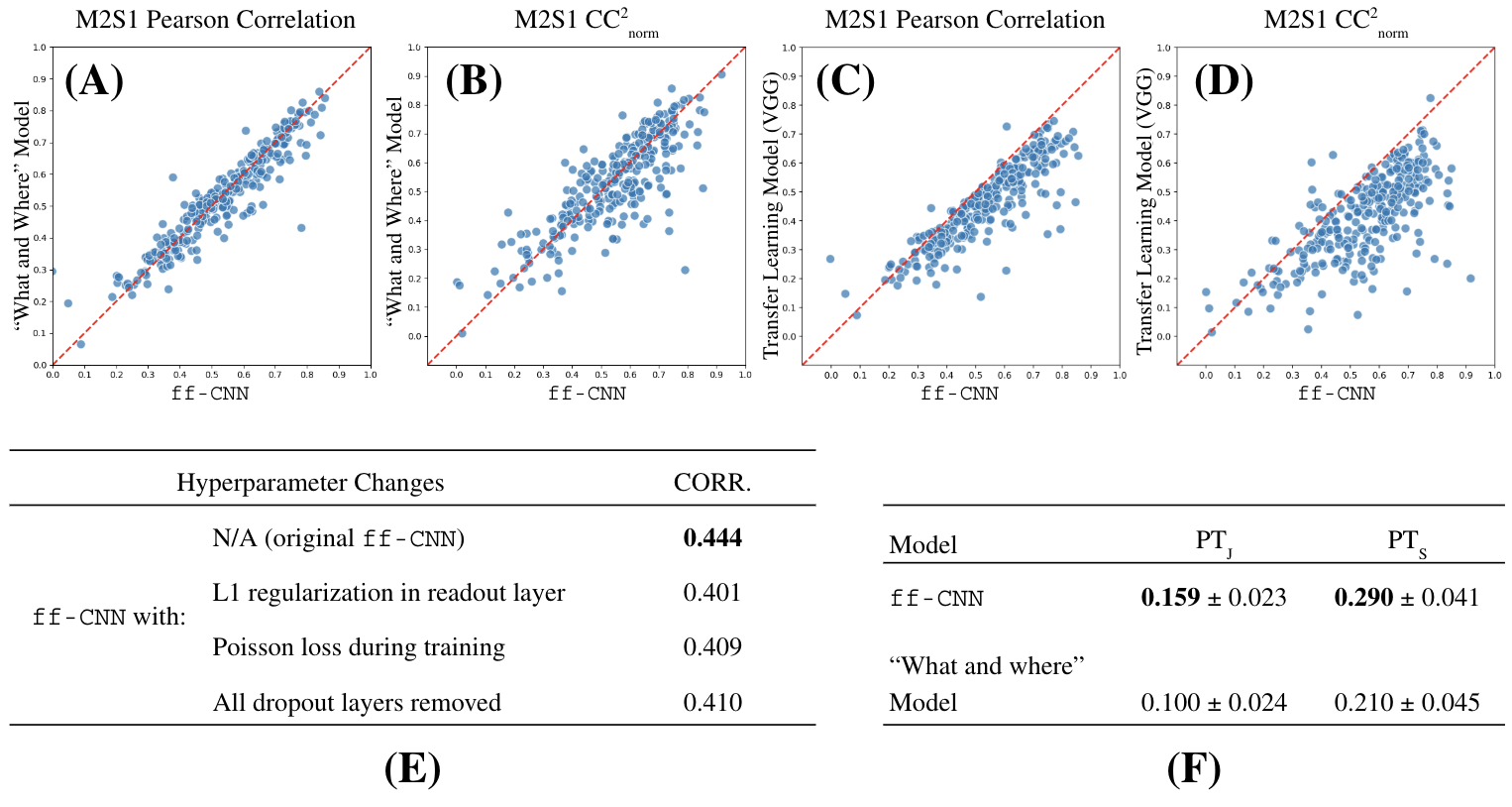}
    \caption{Comparing the performance of \texttt{ff-CNN}. \textbf{(A), (C)} shows the Pearson correlation comparisons of \texttt{ff-CNN} with the "what and where" (shared core factorized) model and the transfer learning (goal driven VGG) model, respectively. \textbf{(B), (D)} show CC norm squared comparisons between models. \textbf{(E)} shows hyperparameter experiments on a subset of 50 neurons from M2S1. \textbf{(F)} Comparison between \texttt{ff-CNN} and the "what and where" model on the ability to capture  M2S1 tuning peaks. All results on M1S1 are similar.}
    \label{comp_model}
\end{figure}
Four major classes of models are found in neural response prediction literature: (1) transfer learning models, (2) single CNN feedforward models, the (3) shared core factorized (“what and where”) model \citep{klindt2018neural, lurz2021generalization}, and more recently a (4) transformer based model for mouse V1. The first three types of models have been used in macaque V1, whereas the transformer based model has only been used in mouse V1. We experimented with the first three classes of models and found that for our dataset, the performance of our baseline feedforward models (\texttt{ff-CNN}) are comparable, and at times better, to the transfer learning model and the shared core factorized mode (see Figure \ref{comp_model}). Thus, we used the single CNN feedforward model as our baseline model. We did not compare our model with transformer based models such as ViT \citep{li2023v1tlargescalemousev1}, which use deep and complex layers to achieve SOTA neural response prediction. The focus of our research and our contributions are different. We demonstrate that surround contextual modulation is critical in predicting the peak responses of macaque V1 neurons in response to natural images. Towards this, we used self-attention to model horizontal interactions, rather than just using an entire transformer module to achieve SOTA performance.

\newpage
\subsection{Comparisons to Other Established Metrics}
\label{a_comp_metric}
\label{A.1}
\begin{figure} [h]
    \centering
    \includegraphics[width=1\linewidth]{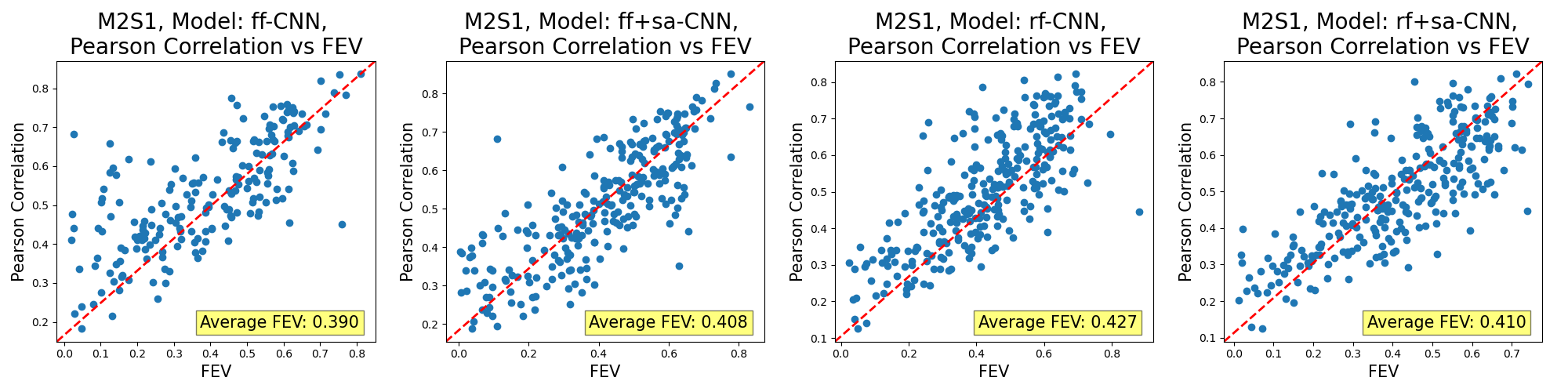}
    \caption{Pearson correlation versus FEV scatter plot for different models. Average FEV is calculated by averaging over neurons in M2S1. Note that the performance of models improve relative to the \texttt{ff-CNN} baseline when evaluated based on FEV, following a similar trend to that of Pearson correlation.}
    \label{fev}
\end{figure}

\begin{figure} [h]
    \centering
    \includegraphics[width=1\linewidth]{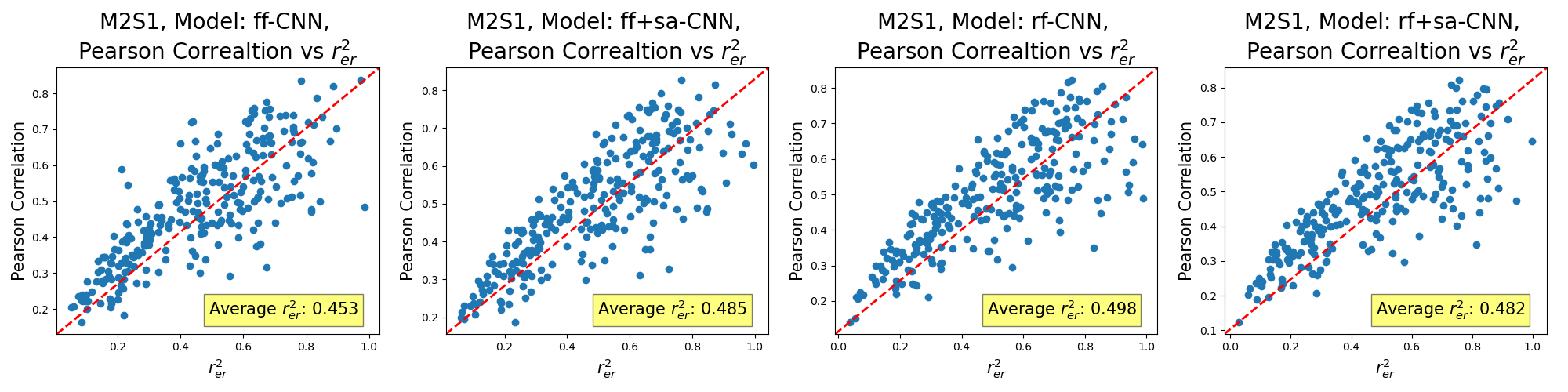}
    \caption{Pearson correlation versus $r^2_{er}$  scatter plot for different models. Average $r^2_{er}$  is calculated by averaging over neurons in M2S1. Note that the performance of models improve relative to the \texttt{ff-CNN} baseline when evaluated based on $r^2_{er}$, following a similar trend to that of Pearson correlation.}
    \label{r2er}
\end{figure}
We compute measures FEV and $r^2_{er}$ for each of our models. We calculated the average FEV and $r^2_{er}$ across all neurons in M2S1 for different models. Similar to Pearson correlation and $CC^2_{norm}$, FEV and $r^2_{er}$ measures are similar across models used in our study, and are comparable to established baselines (see Figure \ref{fev} and Figure \ref{r2er}). Despite FEV and $r^2_{er}$ taking prediction magnitudes into account, they are still unable to capture the peak tuning properties. This can be attributed to the high degree of sparsity in individual neurons’ tuning curve (to natural stimuli), where only 0.4\% of the stimuli above half height on average. Because standard measures of performance are heavily influenced by remaining 99\% low-responding stimuli, they are not sufficient for capturing the sharp tuning curves peaks we observed (i.e.recognizing the most preferred stimuli of the neurons).

\newpage
\subsection{FCL decomposition: average contribution from each hypercolumn}
\label{a_fcl_decomp}
\begin{figure}[H]
    \centering
    \includegraphics[width=0.329\linewidth]{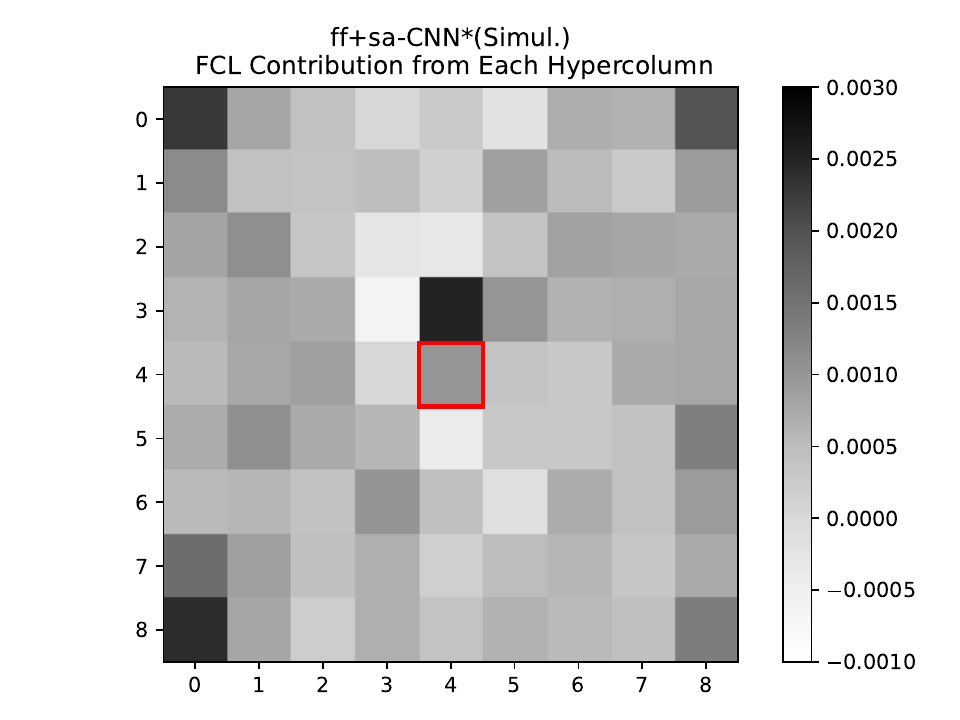}
    \includegraphics[width=0.329\linewidth]{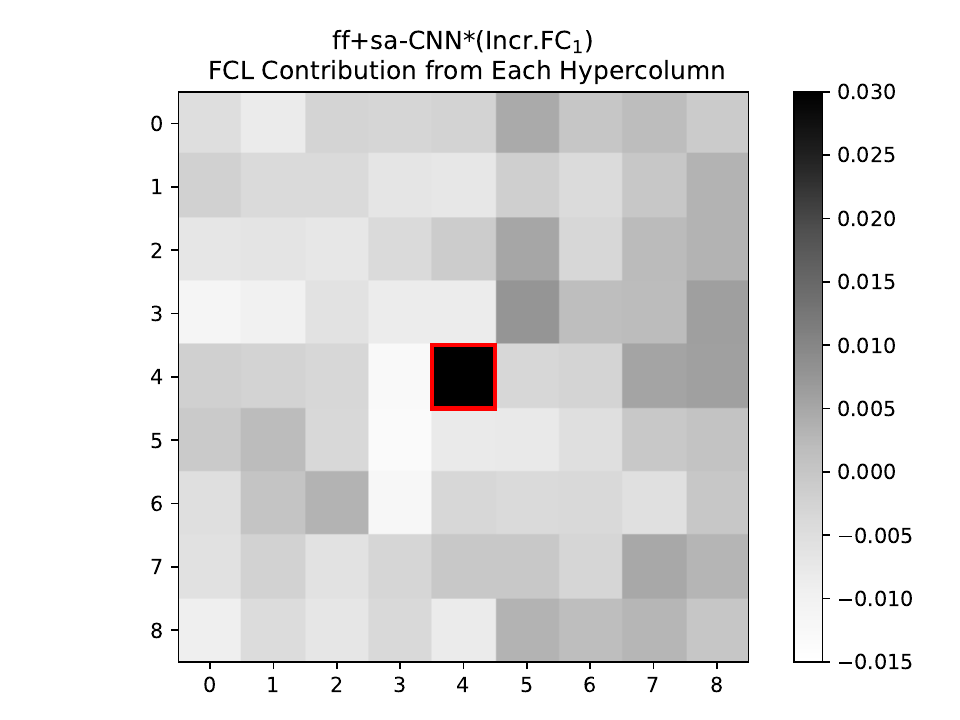}
    \includegraphics[width=0.329\linewidth]{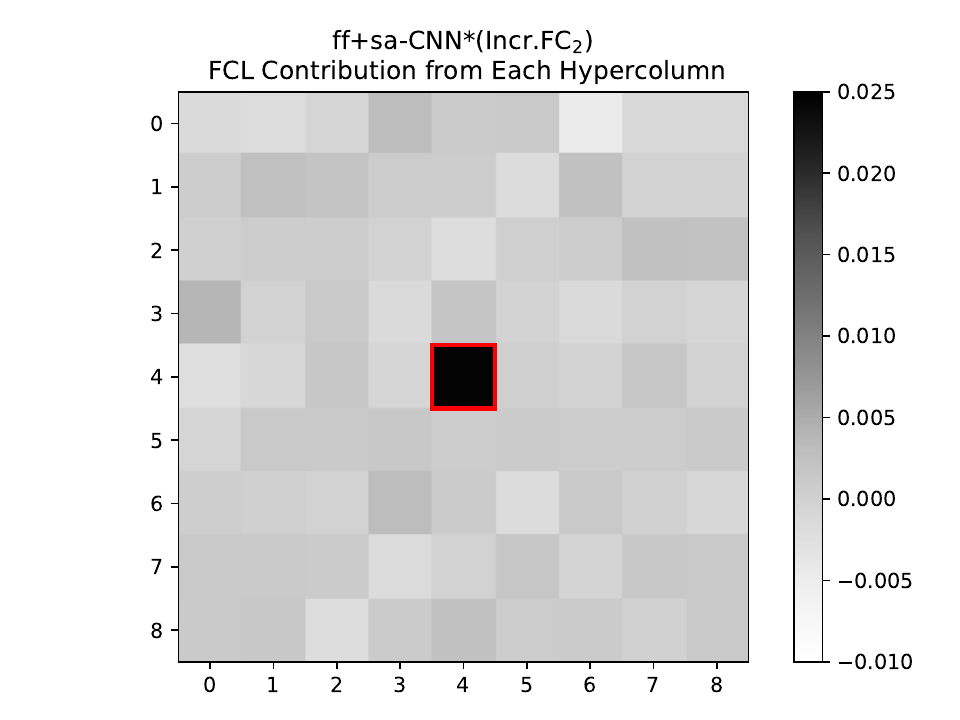}
    \caption{Average FCL contribution from each hypercolumn for M1S1. Center hypercolumn outlined in red. Each heatmap is independently contrast normalized. Note that the distribution in the first heatmap is evenly distributed compared to the others (see contrast scale).}
    \label{fclhm}
\end{figure}

Figure \ref{fcld} plots the individual contribution of the center hypercolumn with the sum of all surrounding hypercolumns. We observed an evenly distributed contribution from all hypercolumns in the simultaneously trained model, but a strong center contribution in incrementally learned models. This effect is further observed when we display the average contribution of each hypercolumn (see Figure \ref{fclhm}).

\subsection{Models that can capture peak tuning exhibit interpretable contextual modulation such as association fields in the self-attention module}
\label{a_att}
\begin{figure}[H]
    \centering
    \includegraphics[width=0.325\linewidth]{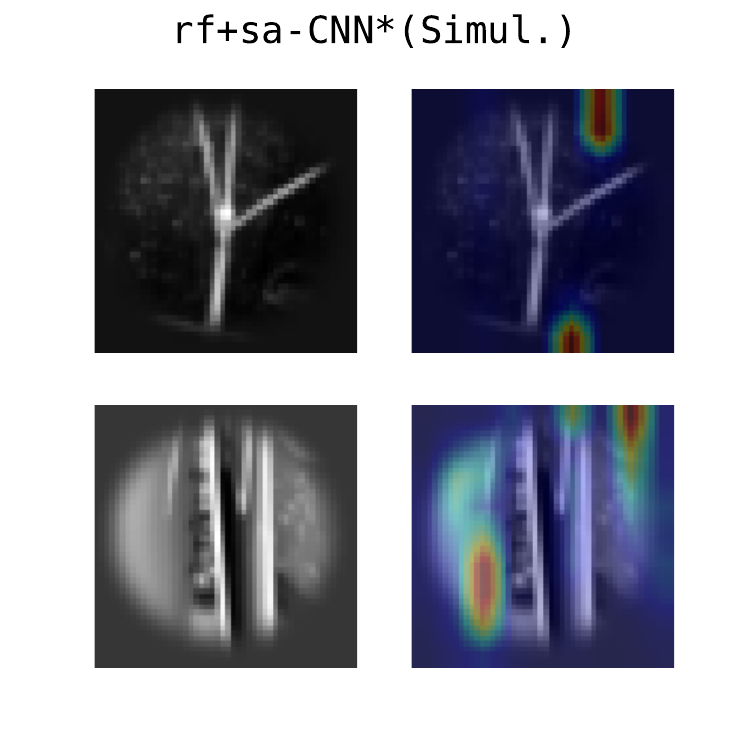}
    \includegraphics[width=0.325\linewidth]{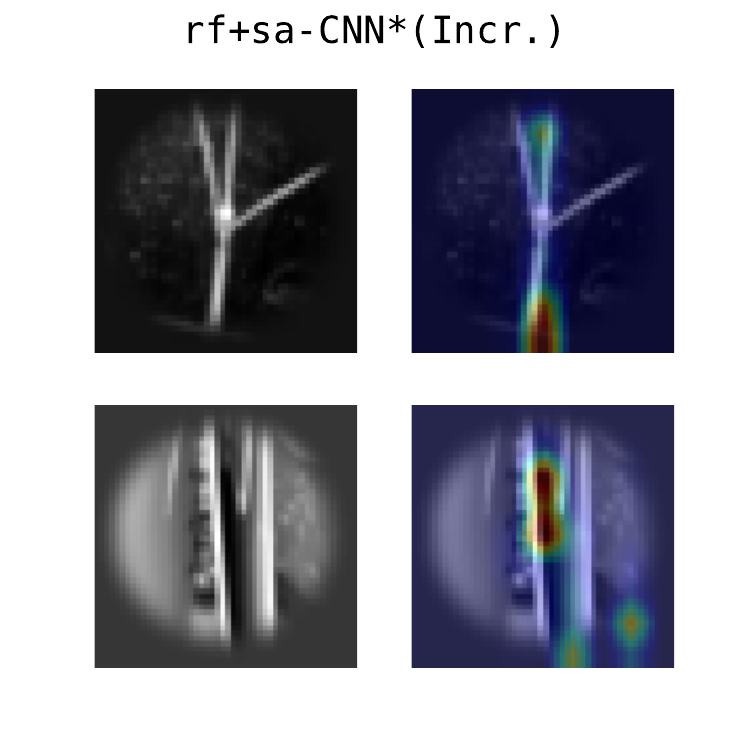}
    \includegraphics[width=0.325\linewidth]{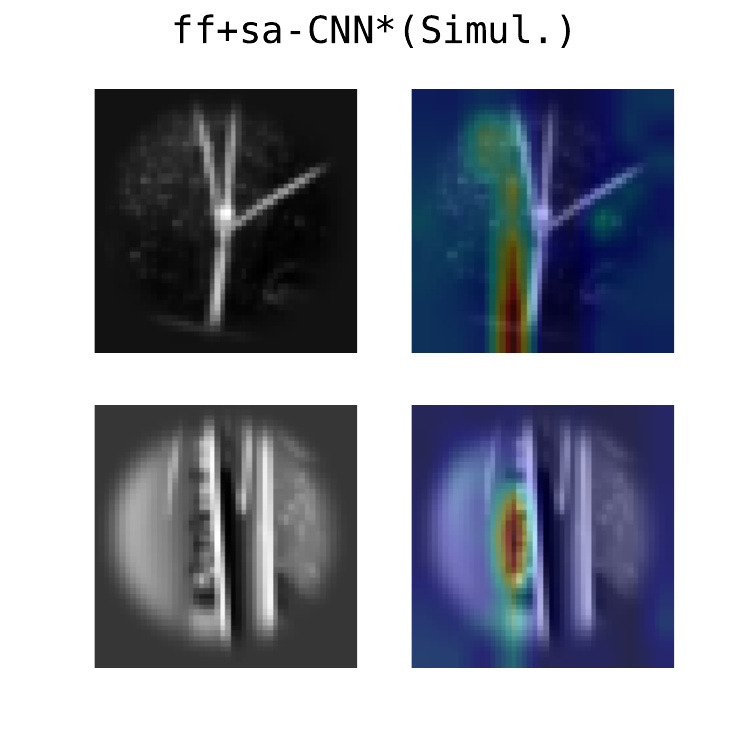}
    \caption{Attention highlighting for incremental learning models. Top two highest response inducing image are shown for M1S1 neuron 153. Note that \texttt{rf+sa-CNN$^*$(Incr)}, \texttt{ff+sa-CNN$^*$(Incr.FC$_1$)}, and \texttt{ff+sa-CNN$^*$(Incr.FC$_2$)} all have the same attention map due to the freezing scheme. The center hypercolumn is queried for highlighted images.}
    \label{hlight}
\end{figure}

Comparing the attention highlighting between self-attention models from the incremental learning experiment, we observe that models learned incrementally have a more focused attention versus equivalent-architecture counterparts trained simultaneously (as shown in Figure \ref{hlight}). Models with strong peak tuning, which incorporate the surround properly,  displays association field effects, focusing on similar patterns as present in the receptor field. Furthermore, because of the incremental freezing scheme between all incremental models, they have the same attention despite variations in the readout layer. However, incremental models display a focused attention, meaning the initially trained SA representation using the CTL (in \texttt{rf+sa-CNN$^*$(Incr)}) allows for proper learning of attention weights. This further supports that the $30$ to $1$ paramter bottleneck in the CTL is not a limiting factor, and the the gain in performance in the latter incremental models (in \texttt{ff+sa-CNN$^*$(Incr.FC$_1$)} and \texttt{ff+sa-CNN$^*$(Incr.FC$_2$)}) are associated with the FCL. 

\newpage
\subsection{Self-attention CNNs are data efficient}
\label{a_de}
\begin{table}[H]
  \caption{Average Pearson correlation and peak tuning metrics for models trained at different data sizes of M1S1. Correlation SEM $=0.008$ was consistent across models. \\}
  \centering
  \footnotesize  %
  \label{def}
  \begin{tabular}{lcccccc}
    \toprule
    & \multicolumn{3}{c}{25\% of M1S1} & \multicolumn{3}{c}{50\% of M1S1} \\
    \cmidrule(lr){2-4}\cmidrule(lr){5-7}
    Model & $\Delta$ CORR. of \texttt{ff-CNN} & PT$_J$ &PT$_S$  & $\Delta$CORR. of \texttt{ff-CNN}  & PT$_J$ &PT$_S$  \\
    \midrule
    \texttt{ff-CNN}     & $0.0\%$   & $0.232$  & $1.026$ & $0.0\%$ & $0.762$ & $1.556$ \\
    \texttt{ff+sa-CNN} & $+7.2\%$   & $0.927$ & $1.656$ & $+7.6\%$ &  $1.026$& $2.715$ \\
    \texttt{rf-CNN}    & $+26.9\%$  & $0.000$ & $0.000$ & $+21.1\%$ & $0.000$ & $0.000$\\
    \texttt{rf+sa-CNN} & $+23.6\%$   & $0.000$ & $0.000$ & $+19.8\%$ & $0.000$ & $0.000$\\
    \bottomrule
  \end{tabular}
\end{table}

We trained baseline feedforward CNN models and their counterparts with self-attention at various training dataset sizes. We conclude that percentage improvements over \texttt{ff-CNN} are furthered at lower data constraints (as shown in Table \ref{def}), alluding to the potential efficiency of SA in accumulating surround information compared to other context mechanisms. We note that \texttt{rf-CNN} is the most data efficient when evaluated solely on Pearson correlation. However, it is important to see from Table 3 that at $25\%$ and $50\% $data, \texttt{rf-CNN} and \texttt{rf+sa-CNN} completely fail in the peak tuning index, indicating that the models were entirely unable to model the response magnitude of the highest excitatory images. This lends to our claim that although \texttt{rf-CNN} does well in correlation, contextual information (as is present in \texttt{ff-CNN} and \texttt{ff+sa-CNN}) is necessary to capture peak responses.

\subsection{Population tuning curves}
\label{a_pop_tc}
\begin{figure}[H]
    \centering
    \includegraphics[width=1\linewidth]{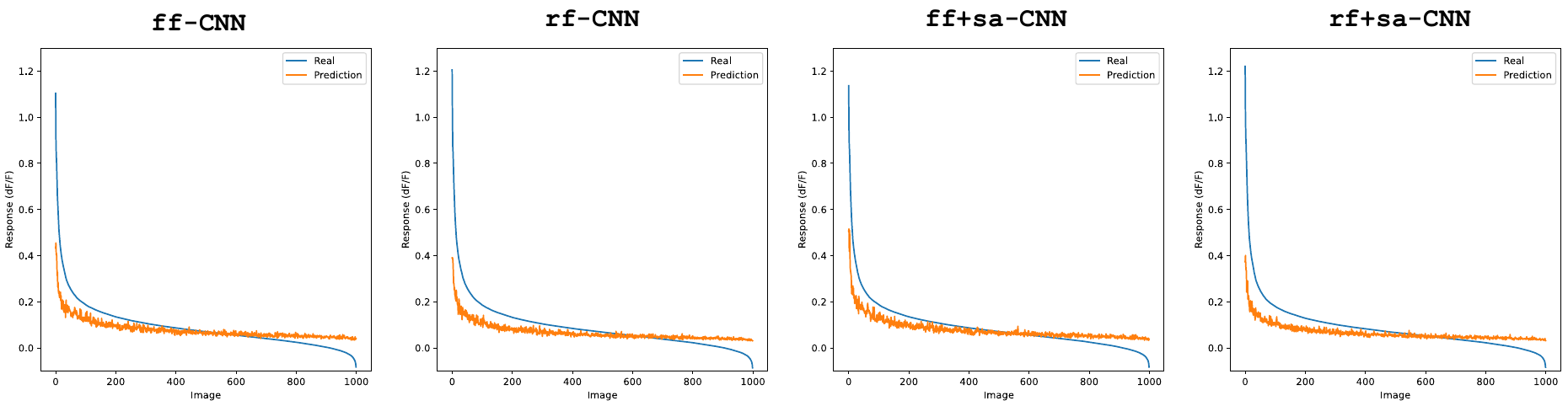}
    \caption{Population average tuning curves for M1S1.}
    \label{poptc}
\end{figure}
Differences in peak tuning can also be observed in the population tuning curves (see Figure \ref{poptc}). Average curves are derived by calculating rank ordered tuning curves for each neuron individually, then averaging over the image number across neurons.

\subsection{V mapping parameter $\gamma$ in self-attention}
\label{a_vmap}
In the self-attention layer (as shown in Figure \ref{z_arch}), the V mapping parameter $\gamma$ allows further factorization of inter-channel mixing and spatial interactions. Toggling $\gamma =$ False removes the transformed value vector, and attention weights instead directly on the input representation. Note that the $\gamma$ parameter does make a difference performance wise. The only difference between \texttt{rf+sa-CNN$^*$} and \texttt{rf+sa-CNN} is the presence of a SA with $\gamma$ = True block in the former and a SA with $\gamma$ = False block in the latter. \texttt{rf+sa-CNN$^*$} has better correlation, PT$_J$, and PT$_S$ values, meaning allowing for the V mapping in SA allows for more flexibility, despite the lack of a $3\times 3$ convolution and FCL layer in these models.

\newpage
\subsection{CTL channel number bottleneck}
\label{a_ctl_chan}
In models with a CTL readout, the final layer is performing a $30$ to $1$ or $32$ to $1$ weighted sum, depending on the number of channels. Thus, an issue we considered was that such a narrow final layer would inhibit proper backpropagation of error signals to upstream modules. To address the concern of a $30 \rightarrow 1$ mapping in the CTL layer being too tight of a an initial bottleneck, we trained \texttt{rf+sa-CNN} with $c=375$ channels, so that it would have a $375 \rightarrow 1$ CTL mapping instead. The results were comparable to the \texttt{rf+sa-CNN} with $c=30$, meaning the drop in performance from \texttt{rf-CNN} cannot be attributed to a parameter bottleneck. 

\subsection{Importance of post-self-attention channel mixing} 
\label{a_post_sa_chan}
Additionally, we tested self-attention models without post-SA convolutions (i.e. no $\beta$CPB layers) and observed sharp drops in performances compared to baseline CNNs. This suggests that inter-channel mixing is crucial in processing the output of self-attention into a interpretable representation by the readout layer. We note that transformer block in modern computer vision models employ a multi-layer perceptron immediately after self-attention, which aligns with our findings.

To compare the importance of the $3\times 3$ versus $1\times 1$ kernel size and FCL vs CTL readout as a means of incorporating surround information, we compared the following models:
[$\alpha$CPB $\rightarrow$ $\alpha$CPB $\rightarrow$ SA($\gamma=$ True) $\rightarrow$ $\beta$CPB($k=1$)$\rightarrow$ $\beta$CPB($k=1$) $\rightarrow$ FCL] $\overset{\text{vs}}{\iff}$ [$\alpha$CPB $\rightarrow$ $\alpha$CPB $\rightarrow$ SA($\gamma=$ True) $\rightarrow$ $\beta$CPB($k=3$)$\rightarrow$ $\beta$CPB($k=3$) $\rightarrow$ CTL].

The former with $\beta$CPB($k=1$) and FCL outperformed the latter with $\beta$CPB($k=3$) and CTL. Thus, direct access to all spatial features with a fully connected layer is stronger than convolving the surround into the center. We observe that the FCL is the strongest factor for predicting the peak responses, and is bolstered by the addition SA, as \texttt{ff+sa-CNN} outperforms \texttt{ff-CNN} in peak tuning.

\subsection{Receptive Field Characteristics and Population-Level RF Distribution}
\begin{figure} [h]
    \centering
    \includegraphics[width=0.32\linewidth]{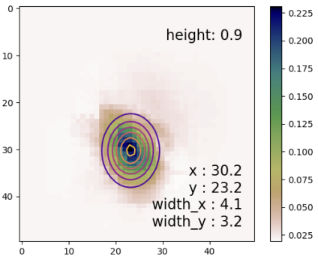}
    \includegraphics[width=0.32\linewidth]{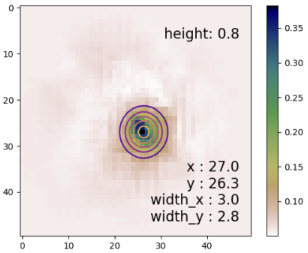}
    \includegraphics[width=0.32\linewidth]{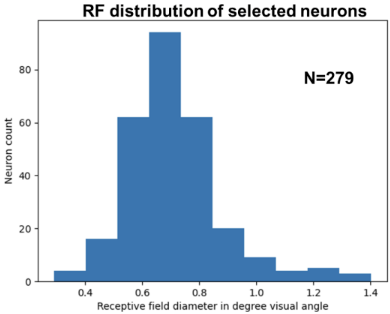}
    \caption{Receptive field properties of neurons and their population-level distribution. \textbf{Left and Center:} two sample neurons' response maps fitted with an elliptical Gaussian. \textbf{Right:} histogram showing the distribution of receptive field diameters (in degree visual angle) for 279 neurons.}
    \label{rebut1}
\end{figure}

We mapped the neural networks’ receptive fields using optimally oriented short bars. The left and center panel of Figure \ref{rebut1} illustrates two example neurons’ receptive fields, indicating that a short bar outside the classical receptive fields does not elicit a response greater than the baseline. The half-height maximum receptive field diameter is approximately 3 pixels (0.4 degree visual angle), corresponding to a 2-STD contour diameter of about 0.8 degree visual angle, consistent with experimentally determined receptive field sizes. The right panel of Figure \ref{rebut1} displays the distribution of mapped receptive fields of neurons for one site.

\subsection{Capturing Modulatory Properties of Extra-Classical Receptive Fields}
To verify that our neural networks exhibit the extra-classical receptive field contextual modulation phenomena noted by the reviewer, we replicated following experiment by \citep{articlecava} on our digital neurons.

The following describes the methodology of the experiment. Neurons were presented with center grating stimuli in optimal orientation and spatial frequency of varying diameters (center-only stimuli), centered on their receptive fields. The diameter of the smallest center-only stimulus eliciting at least 95\% of the neuron’s maximum response defines the GSF (grating summation field). Each neuron was also presented with surround grating stimuli (with gray apertures centered on the receptive field) of varying sizes. The classical receptive field was estimated as the aperture diameter at which the annular stimulus response is significantly below the neuron's maximum response to a circular grating patch. This aperture, referred to as the AMRF, represents where surround-only stimuli do not elicit significant responses above the baseline response. When the center-only stimulus is significantly larger than the size of the classical receptive field, surround suppression—a characteristic example of extra-classical receptive field modulation—is observed.

\begin{figure} [h]
    \centering
    \includegraphics[width=1\linewidth]{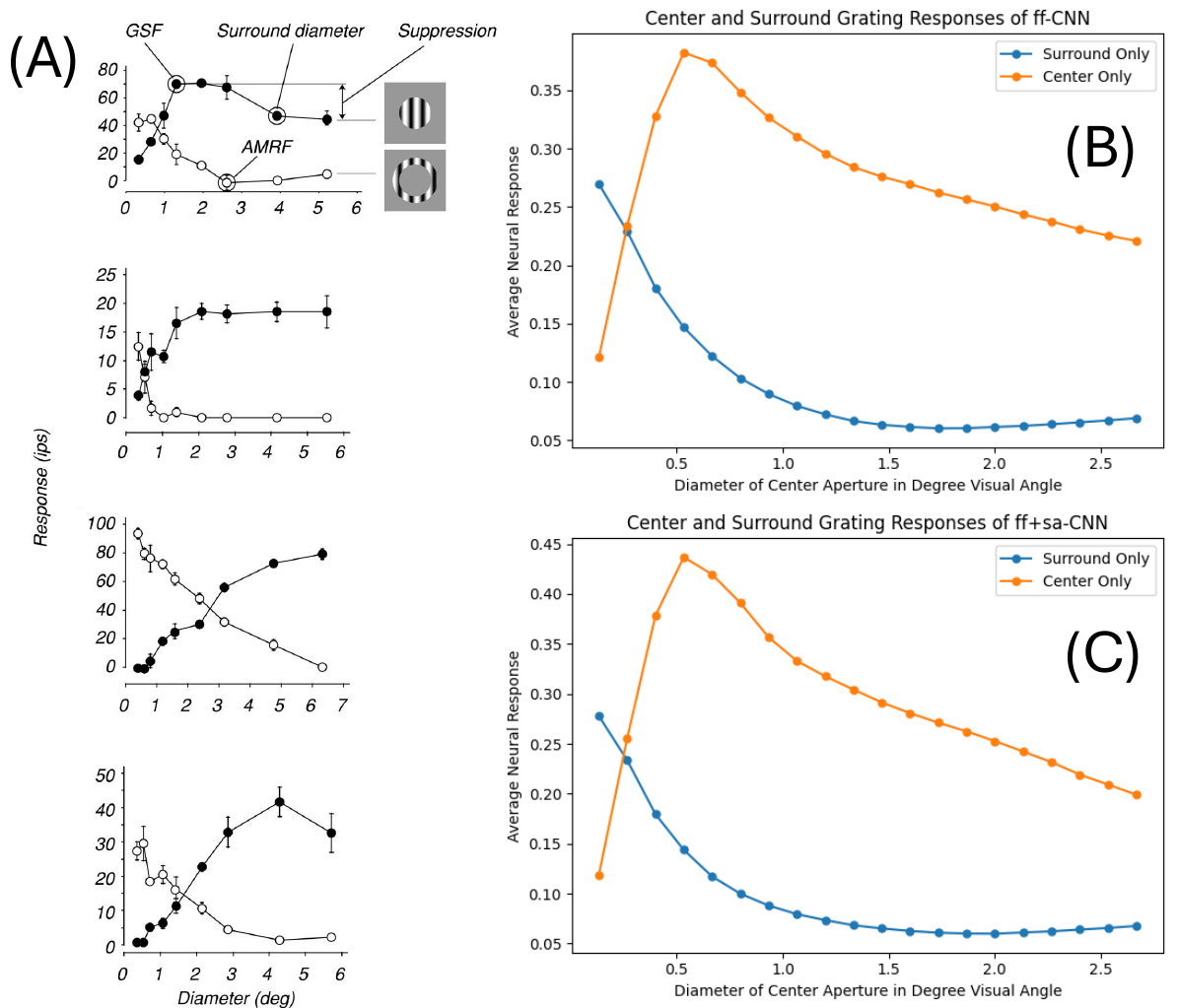}
    \caption{Capturing basic modulatory nature of extra-classical receptive fields. \textbf{(A)} shows four example neurons from \citep{articlecava} study demonstrating the classical surround suppression phenomena. \textbf{(B), (C)} shows the average population response to the center only stimuli and the surround only stimuli for two models, \texttt{ff-CNN} and \texttt{ff+sa-CNN}.}
    \label{rebut2}
\end{figure}

Figure \ref{rebut2}A shows four example neurons from Cavanaugh et al.’s study demonstrating the classical surround suppression phenomena. Figure \ref{rebut2}B and \ref{rebut2}C shows our replication of their experiment, comparing results for the baseline model (ff-CNN) and the self attention model (ff+sa-CNN). The classical receptive field size of our neurons estimated using AMRF was approximately 0.8 degree visual angle. We found that when the center-only stimulus exceeded this classical receptive field size, surround suppression—a hallmark of extra-classical receptive field modulation—was consistently observed.

\newpage
\subsection{Performance comparison of different models at varying data sizes}
\begin{figure} [h]
    \centering
    \includegraphics[width=0.65\linewidth]{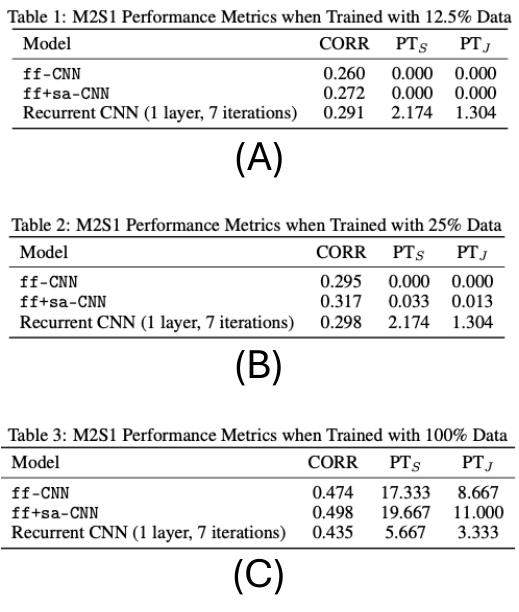}
    \caption{Performance of \texttt{ff-CNN}, \texttt{ff+sa-CNN}, and recurrent CNN at varying data sizes. \textbf{(A)} shows performance metrics at 12.5\% data. \textbf{(B)} shows performance metrics at 25\% data. \textbf{(C)} shows performance metrics at 100\% data}
\end{figure}
Our results indicate that when the full training dataset (100\% data) is utilized, the feedforward CNN augmented with self-attention (\texttt{ff+SA-CNN}) outperforms the recurrently augmented CNN across both overall tuning metrics and peak tuning metrics. However, consistent with the findings of \cite{zhang2022recurrentnetworksimproveneural}, the recurrent augmented CNN exhibits superior data efficiency. Specifically, when the training dataset size is reduced to 25\% or less, the recurrent augmented CNN surpasses both \texttt{ff-CNN} and the \texttt{ff+sa-CNN} in performance.

\subsection{Code for experiments}
\label{a_code}
The code is hosted at the github repository: 
https://github.com/lucanren/sacnn

\end{document}